\documentclass{article}
\usepackage{float}
\usepackage{geometry}
\usepackage{url}
\usepackage{booktabs}
\usepackage[caption=false,font=normalsize,labelfont=sf,textfont=sf]{subfig}
\usepackage[numbers]{natbib}
\usepackage{graphicx} 
\usepackage{array}
\usepackage{afterpage}
\usepackage{xcolor}
\usepackage{listings}
\usepackage{hyperref}
\hypersetup{
    colorlinks,
    citecolor=blue,
    filecolor=blue,
    linkcolor=blue,
    urlcolor=blue
}

\title{Temporal Analysis of NetFlow Datasets for\\ Network Intrusion Detection Systems}
\author{Majed Luay, Siamak Layeghy, Seyedehfaezeh Hosseininoorbin,\\ Mohanad Sarhan, Nour Moustafa, Marius Portmann}
\date{2025}

\begin{document}

\maketitle

\begin{abstract}
%
%
%
%

This paper investigates the temporal analysis of NetFlow datasets for machine learning (ML)-based network intrusion detection systems (NIDS).
Although many previous studies have highlighted the critical role of temporal features, such as inter-packet arrival time and flow length/duration, in NIDS, the currently available NetFlow datasets for NIDS lack these temporal features.
This study addresses this gap by creating and making publicly available a set of NetFlow datasets that incorporate these temporal features~\citep{uq_nids_datasets}.
With these temporal features, we provide a comprehensive temporal analysis of NetFlow datasets by examining the distribution of various features over time and presenting time-series representations of NetFlow features. This temporal analysis has not been previously provided in the existing literature.
We also borrowed an idea from signal processing, time frequency analysis, and tested it to see how different the time frequency signal presentations (TFSPs) are for various attacks. The results indicate that many attacks have unique patterns, which could help ML models to identify them more easily.

\end{abstract}


\section{Introduction}


Maintaining the security and integrity of network infrastructures has become increasingly challenging due to the constantly evolving nature of cyber threats and the vast scale and complexity of modern networks. 
A critical component of network security is monitoring traffic, which provides essential information on potential threats, anomalies, and vulnerabilities. 
However, the overwhelming volume of network traffic has made traditional packet inspection impractical, demanding immense processing power and storage resources while simultaneously raising significant privacy concerns~\citep{WhyFlowsDiffere}. 
A practical solution adopted by many organisations to address these challenges is to implement flow-based network monitoring~\citep{Flow_Based_Literature}. 
This approach aggregates traffic into summarised flows, capturing key communication patterns between endpoints, allowing for efficient analysis, reduced resource demands, and improved privacy protection while still enabling robust threat detection and network management~\citep{NIDS_Datasets_survey}.

Network Intrusion Detection Systems (NIDS) are a vital component of the network security ecosystem, providing real-time monitoring and analysis of network traffic to identify suspicious activities, unauthorised access attempts, and potential security breaches~\citep{Khumar2021ReviewNIDSTrends}.
NIDSs are commonly classified into two main types: signature-based and anomaly-based systems~\citep{NIDS_Literature}.
Signature-based NIDSs rely on databases of known attack signatures, requiring regular updates~\citep{roesch1999snort}. They achieve high accuracy for recognised attacks but face challenges with their variations, polymorphic malware, and zero-day exploits~\citep{zero_day_Attacks}.
In contrast, anomaly-based NIDSs utilise advanced algorithms to learn from traffic patterns, enabling them to adapt to emerging threats and detect anomalies that deviate from normal behaviour~\citep{NIDS_Anomaly_Review}.
To enhance detection capabilities, many modern NIDSs integrate machine learning (ML) techniques, improving both anomaly-based and hybrid approaches~\citep{NIDS_Literature, TPUNIDS}.
The integration of trained ML models into NIDS is referred to as ML-based NIDS~\citep{crossEvaluationML-NIDS}.

ML-based NIDSs are trained to learn patterns in network traffic and enhance anomaly detection by distinguishing between normal and malicious behaviour~\citep{AI_In_Cyber, manocchio2024flowtransformer}.
However, their effectiveness heavily depends on the quality and relevance of the datasets used for training and evaluation~\citep{Dataset_Evaluation}.
In this context, flow-based network monitoring provides a practical solution by summarising traffic into flows, offering a structured representation of network activity that facilitates both training and real-time anomaly detection.
Yet, a significant challenge in using current flow-based benchmark datasets lies in their inconsistent feature sets, which hinder uniform analysis across them.
Each dataset typically presents a unique set of features, complicating the task of comparing and evaluating ML models across different datasets~\citep{sok}.
Sarhan et al. addressed this gap by introducing a NetFlow version of four highly cited flow-based benchmark datasets, standardised to a common feature set~\citep{NF1_dataset,NF2_Datasets}. NetFlow is the most widely used format for collecting flow information in real-world production networks~\citep{NetFlow_2004}.

Although these NetFlow datasets~\citep{NF1_dataset,NF2_Datasets} have addressed the gap in standardised feature sets, they lack most temporal NetFlow features. Consequently, they fall short when employing sequential neural network models or leveraging temporal network traffic to identify attacks.
The inclusion of detailed temporal information in NIDS datasets significantly enhances our ability to analyse traffic patterns and detect anomalies associated with different network attacks~\citep{Temporal_Features_Importance}.
This research bridges this gap by introducing a new NetFlow version of four common NIDS benchmark datasets: UNSW-NB15~\citep{UNSW_Dataset_Original}, BoT-IoT~\citep{BoT_Original_dataset}, ToN-IoT~\citep{ToN_Original_dataset}, and CSE-CIC-IDS2018~\citep{CIC2018_Original_dataset}, which incorporate temporal NetFlow features. These new versions are publicly available and can be accessed via~\citep{uq_nids_datasets}. The details of the temporal features and other specifications of these datasets are discussed in Section~\ref{NetFlow Datasets version 3}.

Upon providing these datasets~\citep{uq_nids_datasets}, we investigate their temporal characteristics through multiple analytical approaches. First, we perform a detailed analysis of flow duration distribution to illustrate the temporal patterns associated with each class of network behaviour within the datasets. Similarly, we examine the distribution of inter-arrival times (IAT) to reveal patterns distinctive to each traffic category. Second, we employ time series representations to dynamically track network activities over time. These visualisations effectively highlight specific attack periods alongside normal traffic flow patterns. Then, both numerical and categorical features are visualised within these representations.

Finally, we apply Time-Frequency Distribution (TFD) representation to explore the frequency components of traffic data over time. Inspired by~\citep{harbic, Niloo_Cattle} work in activity recognition, where TFD successfully identified subtle activity patterns~\citep{harbic, Niloo_Cattle}, we hypothesise that network attacks might also exhibit unique TFD signatures. TFD has been actively used in NIDS, where network traffic is transformed into image formats analysed by convolutional neural networks (CNN) for effective attack classification~\citep{Spectogram_Image, Image-based-spec}. Although our initial investigations have not yet yielded definitive results, they suggest promising directions for future research, potentially leading to breakthroughs in how network attacks are detected and classified. 


 By conducting a thorough analysis of the network's behaviour through NetFlow datasets, we lay a foundational understanding of their network dynamics. This step is crucial as it provides insights into the typical traffic patterns and interactions within the network, fostering a human-level understanding of network behaviours. Such insights are instrumental in designing more targeted and effective strategies for network monitoring and anomaly detection, even without directly engaging in the development or evaluation of machine learning models~\citep{UnderstandNetworkBeforeML}. Our main contributions in this work are outlined as follows:

\begin{itemize}
\item Comprehensive Temporal Analysis of Network Traffic:
We conduct an extensive temporal analysis to demonstrate the evolving dynamics of network traffic and security threats. Through detailed visualisations, including traffic distribution patterns, flow length distributions per attack class, and time-frequency domain representations, we provide novel insights into network behaviour, advancing the understanding of temporal aspects in network security.

\item Public Release of NetFlow-Based Datasets with Temporal Features:
We convert four widely used benchmark NIDS datasets into the NetFlow format, incorporating temporal features that were previously absent in available NetFlow-based benchmark datasets. These enhancements standardise the dataset format, ensuring consistency for machine learning model evaluation, and significantly improve their utility in temporal analysis, leading to more accurate anomaly detection. Moreover, we make these enriched NetFlow datasets publicly available, providing a valuable resource for the research community to support ongoing advancements in machine learning based network intrusion detection.
\end{itemize}

The structure of the paper is as follows: Section 2 reviews related work, Section 3 describes the NF3 datasets, Section 4 presents the temporal analysis, and Section 5 concludes the paper with future work directions.

\section{Related Works}

Dataset analysis is essential to understand the strengths and limitations of different NIDS datasets. Recent studies~\citep{ahmed-survey} and~\citep{Mussa_Datasets_analysis} have surveyed and compared publicly available NIDS datasets. These analyses highlight their diverse characteristics and limitations, noting that the quality of a dataset can significantly impact the performance of detection models. For instance, some datasets do not accurately mirror real-world network scenarios, thereby affecting the reliability of the research conducted using them. In one case, the traffic patterns of NetFlow datasets are  directly compared with real-world traffic, identifying significant discrepancies in statistical features between synthetic and actual datasets~\citep{benchmarkingThebenckmark}. However, the comparison overlooks the analysis of malicious flows and does not address the temporal dynamics of network interactions. Similarly, authors in~\citep{input2024complexity} focused on the complexity of inputs between real-world and lab-based traffic but stopped short of extending this analysis to temporal sequences, which are essential for uncovering deeper behavioural insights.

Further, researchers in~\citep{Khumar2021ReviewNIDSTrends} have explored how dataset characteristics influence NIDS performance, underlining the critical role of careful dataset selection. Their citation-based analysis highlights the popularity of various NIDS approaches, guiding future research directions in the field.  Additionally,~\citep{sok} provides a thorough review of methodologies for evaluating NIDS models and stresses the importance of testing and evaluating these models across multiple datasets to ensure their robustness and applicability. Aligning with this recommendation, our work enriches the field by equipping four widely recognised NIDS datasets with standardised NetFlow features.

To elaborate on their role as benchmarks, recent studies have focused on understanding normal traffic patterns in NIDS datasets to enhance anomaly detection capabilities. Studies such as~\citep{mining_Anomalies2005, empirical,structuralanalysis9} attempt to understand the normal traffic at a level that any deviation will be detected as a suspicious threat. The authors in~\citep{mining_Anomalies2005} demonstrated the necessity of monitoring the traffic feature distributions as it can be a good proof of anomalies. In their study, they worked with collected network data with injected anomalies and found that these anomalies fall into distinct clusters. The authors in~\citep{empirical} highlighted the advantages of using entropy-based approaches for anomaly detection. Their investigation focused on both flow header and behavioural features and it demonstrated a strong correlation between entropy values, which offers comparable effectiveness in detecting anomalies.~\citep{structuralanalysis9} proposed a network traffic modelling based on analysing the source-destination flows in a network. 

Another significant body of work concentrates on analysing specific traffic features to gain deeper insights into network behaviours. For example, some works focus on analysing flow length features, as it offers deep insights into network traffic behaviour and is a focal point of extensive research~\citep{FLD_Analysis,FLD_Classify}. The studies in~\citep{elephant1,elephant2,elephant3} focused on elephant flow detection, which refers to the process of identifying large, long-lived network flows that consume a significant amount of bandwidth. Typically, benign traffic exhibits a certain range of flow lengths depending on the application protocols and user behaviour patterns. In contrast, malicious traffic, such as that generated by attacks like port scanning, DoS attacks, or data exfiltration, often shows distinct flow length characteristics that deviate from the norm~\citep{SlowDDoS_RelaredWork}. 

Additionally, a number of studies emphasise the significance of the IAT feature, alongside other crucial flow characteristics, for effective monitoring of traffic patterns~\citep{netwrktrafficanomalydetection11, TheNatureOfDatacenter}. The work in~\citep{netwrktrafficanomalydetection11} analysed the traffic characteristics, including IAT, across ten diverse data centre networks across various administrative domains including universities, enterprises, and cloud service providers. This analysis was aimed at understanding the distinct traffic patterns and the underlying dynamics of these data centres by meticulously examining both flow and packet-level attributes associated with different layer-7 applications. Meanwhile, the authors in~\citep{TheNatureOfDatacenter} extended this analysis by examining the distribution of key traffic features. Their data collection methodology encompassed three levels of network monitoring: SNMP counters for basic metrics, sampled flow or packet header data for more granular insights, and deep packet inspection for detailed content analysis. While the primary focus of the study was on evaluating network traffic volume and identifying congestion, it also covered various other traffic patterns, including server interactions, flow metrics, and bandwidth usage.  Despite the proven benefits of temporal analysis in these fields, NetFlow data has not been extensively explored in this regard.

Regarding the standard flow format like NetFlow~\citep{NetFlow}, temporal analysis remains under-explored. Studies have explored the effectiveness of sequential learning models, such as Long Short-Term Memory (LSTM), in extracting temporal characteristics from NetFlow data for NIDS~\citep{Temporal_NetFlow}. Some researchers adopted the CNN and LSTM models simultaneously to construct a hybrid model~\citep{stidm,stidm2}. CNN is mainly used to extract spatial features and has made many computer vision applications remarkable~\citep{Review_ComputerVision}. In~\citep{stidm,stidm2}, the authors introduce the Spatial and Temporal Aware Intrusion Detection Model (STIDM). STIDM is a spatio-temporal feature extraction model designed to analyse IAT features between consecutive packets. This model employs a well-known CNN architecture, LeNet-5, for extracting spatial features, complemented by a modified LSTM to capture temporal patterns. While this method allows for grouping packets into flows, it does not effectively facilitate the determination of broader temporal patterns across NetFlow data, making the exploration of temporal dependencies at the NetFlow level unfeasible. 

The authors in~\citep{Temporal_NetFlow} explore temporal sequences of network traffic flows that denote patterns of malicious activities. The main focus was not to compete with the state-of-the-art solutions but rather to find specific temporal patterns, if exist, for each attack class. The paper investigates the use of LSTM neural networks to learn temporal patterns in network flows for NIDS and compares the performance of the LSTM to a static Feed-forward Neural Network (FNN) model. Their goal is similar to ours but we are more interested in understanding the temporal aspect at the feature level within NetFlow datasets.

Building on these initial forays into temporal NetFlow analysis, our research aims to provide a deep understanding of the temporal features in NetFlow datasets. We specifically focus on the temporal dynamics of these datasets without the direct intention of developing new anomaly detection models. Instead, our objective is to enrich the analytical tools available for network security, providing insights that are crucial for the real-time detection and analysis of network anomalies. By making these enriched datasets publicly available, we also contribute to the broader research community, offering resources that enable more detailed and effective analysis of network behaviours.


\section{NIDS Datasets}
High-quality datasets are essential for the effective evaluation and development of ML-NIDS systems~\citep{Dataset_Evaluation}. Historical datasets such as KDD Cup 99 and NSL-KDD, while once foundational, have become less relevant due to their outdated attack patterns from the late 1990s and early 2000s~\citep{KDD_limitations}. The evolving nature of cyber threats highlights the necessity for up-to-date datasets that mirror current network environments and attack patterns~\citep{UNSW_Dataset_Original}. This ensures that ML models are evaluated against current challenges and tailored to address emerging cybersecurity threats, enhancing their effectiveness and relevance. This paper uses four contemporary datasets for this purpose, each providing a rich source of network traffic data reflecting current network environments: \begin{itemize} \item \textbf{UNSW-NB15~\citep{UNSW_Dataset_Original}:} Developed by the Cyber Range Lab of the Australian Centre for Cyber Security (ACCS) using the IXIA PerfectStorm tool to create a mix of normal and malicious traffic, including 12 synthetic attack scenarios. \item \textbf{BoT-IoT~\citep{BoT_Original_dataset}:} Also created by ACCS, this dataset includes a comprehensive mix of benign and malicious traffic covering five types of attack scenarios. \item \textbf{ToN-IoT~\citep{ToN_Original_dataset}:} A heterogeneous dataset encompassing telemetry data of IoT services and operating system logs, designed to assist in the development and evaluation of NIDSs. This dataset was also created by ACCS and it contains 9 attack classes. \item \textbf{CSE-CIC-IDS2018~\citep{CIC2018_Original_dataset}:} Released by a collaboration between the Communications Security Establishment (CSE) and the Canadian Institute for Cybersecurity (CIC), this dataset focuses on simulating realistic network traffic combined with non-overlapping attacks. \end{itemize}

Despite their utility for single dataset evaluation, the inconsistency in feature sets across various datasets makes it challenging to ensure fair and reliable evaluations of ML-NIDS models~\citep{sok}. To address this gap, previous efforts have standardised these datasets to a unified NetFlow format~\citep{NF1_dataset,NF2_Datasets}, enhancing their usability for consistent model evaluation. The authors identified 43 features that were most effective in classifying attack classes in the datasets. Table~\ref{nf} shows the full set of features used in the last NetFlow datasets~\citep{NF2_Datasets} and also the missing features proposed in this version (in bold), which will be explained in the next section. 

\begin{table*}[!t]\scriptsize
\centering
\caption{List of the proposed standard NetFlow features and the added temporal features}
\label{nf}
\begin{tabular}{|l|l|}
\hline
\multicolumn{1}{|c|}{\textbf{Feature}} & \multicolumn{1}{c|}{\textbf{Description}} \\ \hline
IPV4\_SRC\_ADDR                        & IPv4 source address                       \\ \hline
IPV4\_DST\_ADDR                        & IPv4 destination address                  \\ \hline
L4\_SRC\_PORT                          & IPv4 source port number                   \\ \hline
L4\_DST\_PORT                          & IPv4 destination port number              \\ \hline
PROTOCOL                               & IP protocol identifier byte               \\ \hline
L7\_PROTO                              & Application protocol (numeric)                 \\ \hline
IN\_BYTES                              & Incoming number of bytes                  \\ \hline
OUT\_BYTES                             & Outgoing number of bytes                  \\ \hline
IN\_PKTS                               & Incoming number of packets                \\ \hline
OUT\_PKTS                              & Outgoing number of packets                \\
\hline
FLOW\_DURATION\_MILLISECONDS           & Flow duration in milliseconds             \\ \hline
TCP\_FLAGS                             & Cumulative of all TCP flags               \\ \hline
CLIENT\_TCP\_FLAGS                              & Cumulative of all client TCP flags                  \\ \hline
SERVER\_TCP\_FLAGS                            & Cumulative of all server TCP flags                  \\ \hline

DURATION\_IN                        & Client to Server stream duration (msec)                      \\ \hline
DURATION\_OUT                      & Client to Server stream duration (msec)                 \\ \hline
MIN\_TTL                          & Min flow TTL                   \\ \hline
MAX\_TTL                        & Max flow TTL \\ \hline
LONGEST\_FLOW\_PKT                               & Longest packet (bytes) of the flow               \\ \hline
SHORTEST\_FLOW\_PKT                            & Shortest packet (bytes) of the flow              \\ \hline
MIN\_IP\_PKT\_LEN                             & Len of the smallest flow IP packet observed                \\ \hline
MAX\_IP\_PKT\_LEN          & Len of the largest flow IP packet observed             \\ \hline
SRC\_TO\_DST\_SECOND\_BYTES                             & Src to dst Bytes/sec              \\ \hline
DST\_TO\_SRC\_SECOND\_BYTES          & Dst to src Bytes/sec        \\ \hline
RETRANSMITTED\_IN\_BYTES                        & Number of retransmitted TCP flow bytes (src-$>$dst)                       \\ \hline
RETRANSMITTED\_IN\_PKTS                     & Number of retransmitted TCP flow packets (src-$>$dst)                  \\ \hline
RETRANSMITTED\_OUT\_BYTES                          & Number of retransmitted TCP flow bytes (dst-$>$src)                   \\ \hline
RETRANSMITTED\_OUT\_PKTS                       & Number of retransmitted TCP flow packets (dst-$>$src)              \\ \hline
SRC\_TO\_DST\_AVG\_THROUGHPUT                               & Src to dst average thpt (bps)               \\ \hline
DST\_TO\_SRC\_AVG\_THROUGHPUT                           & Dst to src average thpt (bps)               \\ \hline
NUM\_PKTS\_UP\_TO\_128\_BYTES                             & Packets whose IP size $<$= 128                \\ \hline
NUM\_PKTS\_128\_TO\_256\_BYTES          & Packets whose IP size $>$ 128 and $<$= 256             \\ \hline
NUM\_PKTS\_256\_TO\_512\_BYTES                        & Packets whose IP size $>$ 256 and $<$= 512                       \\ \hline
NUM\_PKTS\_512\_TO\_1024\_BYTES                     & Packets whose IP size $>$ 512 and $<$= 1024                  \\ \hline
NUM\_PKTS\_1024\_TO\_1514\_BYTES                          & Packets whose IP size $>$  1024 and $<$= 1514                   \\ \hline
TCP\_WIN\_MAX\_IN                       &  Max TCP Window (src-$>$dst)              \\ \hline
TCP\_WIN\_MAX\_OUT                               & Max TCP Window (dst-$>$src)               \\ \hline
ICMP\_TYPE                           & ICMP Type * 256 + ICMP code               \\ \hline
ICMP\_IPV4\_TYPE                             & ICMP Type               \\ \hline
DNS\_QUERY\_ID          & DNS query transaction Id           \\ \hline
DNS\_QUERY\_TYPE                           & DNS query type (e.g., 1=A, 2=NS..)              \\ \hline
DNS\_TTL\_ANSWER                             & TTL of the first A record (if any)               \\ \hline
FTP\_COMMAND\_RET\_CODE          & FTP client command return code         \\ \hline
\textbf{FLOW\_START\_MILLISECONDS} & \textbf{Flow start timestamp in milliseconds} \\ \hline
\textbf{FLOW\_END\_MILLISECONDS} & \textbf{Flow end timestamp in milliseconds} \\ \hline
\textbf{SRC\_TO\_DST\_IAT\_MIN} & \textbf{Minimum IAT (src-$>$dst)} \\ \hline
\textbf{SRC\_TO\_DST\_IAT\_MAX} & \textbf{Maximum IAT (src-$>$dst)} \\ \hline
\textbf{SRC\_TO\_DST\_IAT\_AVG} & \textbf{Average IAT (src-$>$dst)} \\    \hline
\textbf{SRC\_TO\_DST\_IAT\_STDDEV} & \textbf{Standard deviation of IAT (src-$>$dst)} \\    \hline
\textbf{DST\_TO\_SRC\_IAT\_MIN} & \textbf{Minimum IAT (dst-$>$src)} \\    \hline
\textbf{DST\_TO\_SRC\_IAT\_MAX} & \textbf{Maximum IAT (dst-$>$src)} \\    \hline
\textbf{DST\_TO\_SRC\_IAT\_AVG} & \textbf{Average IAT (dst-$>$src)}\\    \hline
\textbf{DST\_TO\_SRC\_IAT\_STDDEV} & \textbf{Standard deviation of IAT (dst-$>$src)} \\    \hline
\end{tabular}
\end{table*}

\section{NetFlow Datasets version 3}\label{NetFlow Datasets version 3}
This section introduces NF3-Datasets, the third iteration of NetFlow-based datasets converted from the four aforementioned datasets~\citep{UNSW_Dataset_Original,CIC2018_Original_dataset,ToN_Original_dataset,BoT_Original_dataset}. These conversions standardise the representation of network flows, enabling consistent cross-dataset analysis and facilitating advanced intrusion detection research. The selection of features extracted from the original datasets was rigorously assessed in the previous version~\citep{SARHAN_SHAP}; consequently, the current datasets retain the established feature set while also enriching them by adding time-related features, as explained below.

\subsection{Temporal Features}

As can be seen in Table~\ref{nf}, the list of features included in this version is the same as the previous version~\citep{NF2_Datasets} plus the temporal features. 
The added features provide a temporal dimension for network traffic analysis, facilitating the precise identification and correlation of events over time. 
The temporal features listed can be classified into two categories: ``Flow Timing" for determining the start and end time of each flow in milliseconds format, and ``Inter-Packet Arrival Time" for including various statistics of the arrival times between consecutive packets in a flow.
 
Flow timing enables researchers to accurately sequence network flows, ensuring that data aggregation and analysis reflect the true dynamics of network interactions.  In the datasets, these timing values are stored in Unix timestamp format, which represents the number of milliseconds elapsed since January 1, 1970 (UTC).
Precise timing is critical for activities such as event correlation, where understanding the order and duration of flows can reveal patterns indicative of coordinated attacks or system anomalies.

Inter-packet Arrival Time (IAT) serves as another crucial metric, offering valuable insights into the dynamics of network traffic. IAT is calculated as the time interval between the arrival of consecutive packets at a network device, either from source to destination or vice versa. To accurately capture this metric, each packet's timestamp is recorded upon arrival, and the difference between consecutive timestamps is computed. These time differences are then used to calculate the minimum, maximum, average, and standard deviation for each flow. Although these metrics originate from packet-level observations, they are aggregated at the flow level to provide a more comprehensive view of traffic patterns.
Through a detailed examination of the IAT over time, we can gain comprehensive insights into the behaviour of traffic flows. Researchers are attracted to these features because they can uncover subtle deviations from normal traffic patterns~\citep{mining_Anomalies2005,netwrktrafficanomalydetection11, TheNatureOfDatacenter}, providing a deeper layer of analysis that enhances the detection of both sophisticated and low-profile network attacks.\\

\begin{figure}[!t]
    \centering
    \scriptsize
    \includegraphics[width=1\linewidth]{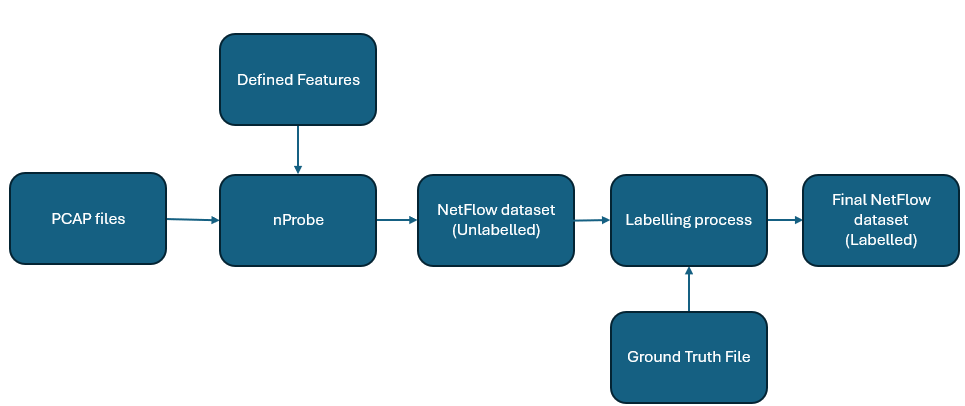}
    \caption{Illustration of the Dataset Conversion and Labeling Process}
    \label{fig:conversion_process}
\end{figure}

\subsection{Conversion Methodology}

The providers of the original datasets~\citep{UNSW_Dataset_Original,CIC2018_Original_dataset,ToN_Original_dataset,BoT_Original_dataset} have released their source files in various formats enabling researchers to adapt and utilise these datasets according to specific research needs and to address known limitations. As seen in~\citep{NF1_dataset,NF2_Datasets}, this flexibility aids in mitigating the feature divergence gap found in NIDS datasets by allowing for the regeneration of datasets with a standardised feature set in NetFlow format.

The process of generating the current version of the NetFlow datasets is the same as previous versions~\citep{NF1_dataset,NF2_Datasets}, displayed in Figure~\ref{fig:conversion_process}. The implementation was conducted on a machine running Ubuntu 20.04 LTS equipped with nProbe software. The nProbe is developed by Ntop~\citep{Ntop}, and is specifically designed to process and convert raw network traffic into the NetFlow records. As can be seen in Figure~\ref{fig:conversion_process}, the workflow initiates with the acquisition of the PCAP files, which are publicly available for each dataset on their respective official websites. Given the extensive volume of data, significant storage capacity is required; for instance, the CSE\_CIC\_IDS2018 dataset~\citep{CIC2018_Original_dataset} alone comprises more than 4,000 PCAP files, totalling over 400 gigabytes. Once collected, the PCAP files undergo conversion through the following nProbe command invocation: \\

\begin{verbatim}
nprobe -i file.pcap -V 9 --dont-reforge-time -T %feature1%feature2%featureN 
--dump-path <path> --dump-format t --csv-separator '#'
\end{verbatim}

In the above command, the \texttt{-i} option specifies the input file, \texttt{-V 9} sets the NetFlow version to 9, and \texttt{--dont-reforge-time} preserves the original timestamps of the network traffic, ensuring the timing data are not modified to match the time of command execution. The \texttt{--dump-path} option defines the directory for the output files, \texttt{--dump-format t} selects the text file format for the output, and \texttt{--csv-separator '\#'} is used to separate the columns with a \texttt{'\#'} in the resulting files. This configuration extracts 57 different flow features using the \texttt{-T} option, organising them according to the specified criteria.
The outputs generated from executing the nProbe command are a series of text files that chronologically catalogue all flow data with precise temporal information. Then, the text files are seamlessly merged and converted into CSV format, facilitating easy reading and efficient organisation of the datasets.

  


By this stage, we have compiled four datasets containing detailed flow information. These datasets are not yet labelled, which means there is no differentiation between normal and malicious flows, nor identification of specific types of attacks within the malicious flows. The subsequent phase involves labelling each flow based on the comparison with the corresponding ground truth file. Labelling is refined by comparing the precise timestamps and 5-tuple identifiers (Source/Destination IP, Source/Destination Ports, Protocol) to accurately match flows with their respective ground truth labels. The purpose of the labelling stage is to augment the datasets with two columns: one for binary classification and another for multi-class classification. In the binary column, a label of 0 signifies a benign flow, while a label of 1 denotes a malicious flow. The summary of binary labelling is depicted in Table~\ref{tab:dataset_summary}. On the other hand, the multi-class classification column encapsulates the specific type of attack, as documented in the ground truth files, allowing for a granular analysis of threat types. Detailed statistics regarding the distribution of attack classes within the datasets are presented in Table~\ref{attacks_statistics}. 

\begin{table*}[!t]
  \centering
  \scriptsize
  \caption{Summary of Malicious and Benign Flows in NF3-Datasets }
  \label{tab:dataset_summary}
  \begin{tabular}{lrrrr}
    \toprule
    \textbf{Dataset} & \textbf{Malicious Flows} & \textbf{Benign Flows} & \textbf{Total Flows} \\
    \midrule
    NF3-UNSW-NB15 & 127,693(5.40\%) & 2,237,731(94.60\%) & 2,365,424 \\
    NF3-CSE-CIC-IDS2018 & 2,600,903(12.93\%) & 17,514,626(87.07\%) & 20,115,529\\
    NF3-ToN-IoT & 10,728,046 (38.98\%) & 16,792,214(61.02\%) & 27,520,260 \\
    NF3-BoT-IoT & 16,881,819(99.7\%) & 51,989(0.3\%) & 16,933,808 \\
    \bottomrule
  \end{tabular}
\end{table*}
 
\begin{table*}[!b]
    \centering
    \scriptsize
    \caption{Statistics of attack types across the datasets, showing the count of flows categorised under each attack and benign class.}
    \label{attacks_statistics}
    \begin{tabular}{lrrrr}
        \hline
        Attack Type & NF3-UNSW-NB15 & NF3-CSE-CIC-IDS2018 & NF3-ToN-IoT & NF3-BoT-IoT \\
        \hline
        Benign &  2,237,731 & 17,514,626 & 16,792,214 & 51,989 \\
        DoS & 5,980 & 302,966 & 203,456 & 8,034,190 \\
        DDoS & — & 1,324,350 & 4,141,256 & 7,150,882 \\
        Reconnaissance & 17,074 & — & — & 1,695,132 \\
        Backdoor & 4,659 & — & 203,384 & — \\
        Fuzzers & 33,816 & — & — & — \\
        Exploits & 42,748 & — & — & — \\
        Analysis & 1,226 & — & — & — \\
        Generic & 19,651 & — & — & — \\
        Shellcode & 2,381 & — & — & — \\
        Worms & 158 & — & — & — \\
        Web Attacks & — & 2,538 & — & — \\
        Infiltration & — & 188,152 & — & — \\
        BoT & — & 207,703 & — & — \\
        BrutForce & — & 575,194 & — & — \\
        Scanning & — & — & 1,358,977 & — \\
        XSS & — & — & 2,834,435 & — \\
        Password & — & — & 1,594,777 & — \\
        Injection & — & — & 381,777 & — \\
        Ransomware & — & — & 3,971 & — \\
        MITM & — & — & 6,013 & — \\
        Theft & — & — & — & 1,615 \\
        \hline
        Total & 2,365,424 & 20,115,529 & 27,520,260 & 16,933,808 \\
        \hline
    \end{tabular}
\end{table*}

The resultant of labelled datasets are the four finalised datasets that we propose in this paper, designated as NF3-UNSW-NB15, NF3-BoT-IoT, NF3-ToN-IoT, and NF3-CSE-CIC-IDS2018. All four datasets share the same feature set, which allows for better evaluation and comparison when implementing and evaluating ML-NIDS models. The inclusion of timestamp information allows for identifying the exact time of the traffic when the original traffic was captured. It is worth mentioning that the timestamps included in the datasets represent the timestamps documented in their respective PCAP files, not the timestamps at which the data was converted to the NetFlow format. This distinction ensures that the temporal integrity of the original network conditions is preserved in the datasets. 
Following this dataset preparation, the next section will delve into the temporal analysis of these datasets. This analysis aims to explore the dynamic patterns and temporal characteristics of the traffic, providing deeper insights into the timing and progression of the recorded network behaviour.

\section{Temporal Analysis}
Gaining a human-level understanding of network traffic is essential before moving on to predictive modelling~\citep{expersknowledge}. By incorporating temporal information into the NetFlow datasets, we can apply various temporal analysis methods to gain deeper insights into network behaviour. 
As mentioned in the related work section, many studies have explored network attack patterns over time~\citep{Temporal_NetFlow}. However, unlike approaches that often aim at classification, this work focuses primarily on the temporal analysis at the feature level within NetFlow datasets. This analysis is not aimed at classifying or predicting specific types of network attacks but rather seeks to deepen our understanding of the inherent temporal characteristics of network features.
In this section, we analyse NetFlow datasets from multiple perspectives, aiming to uncover insights into the dynamics of network traffic. 

\begin{figure}[!b]
    \centering
    \subfloat[\centering][BoT-IoT]
    {\includegraphics[width=0.49\linewidth]{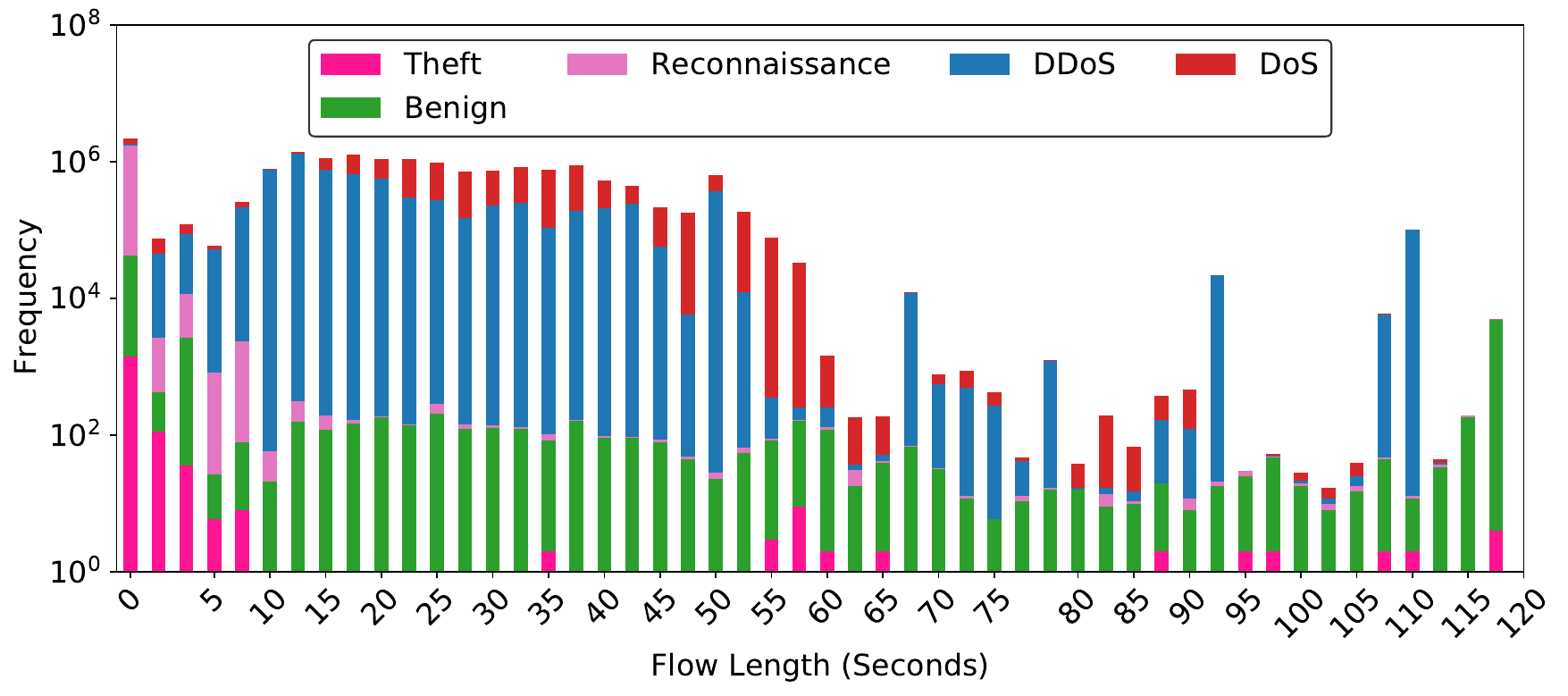}}\hspace{0.01cm}
    \subfloat[\centering][CSE\_CIC\_IDS2018]
    {\includegraphics[width=0.49\linewidth]{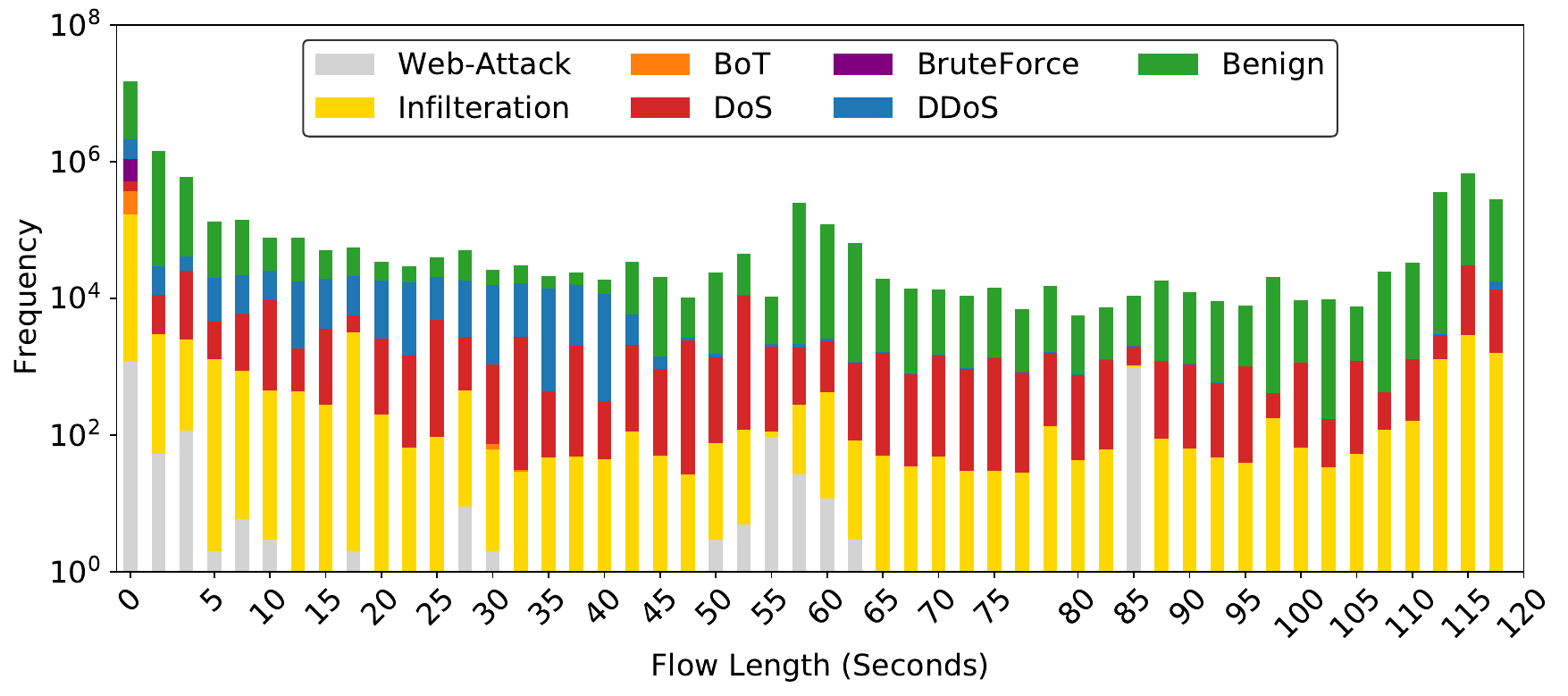}}\hspace{0.01cm}
    
    \subfloat[\centering][ToN-IoT]
    {\includegraphics[width=0.49\linewidth]{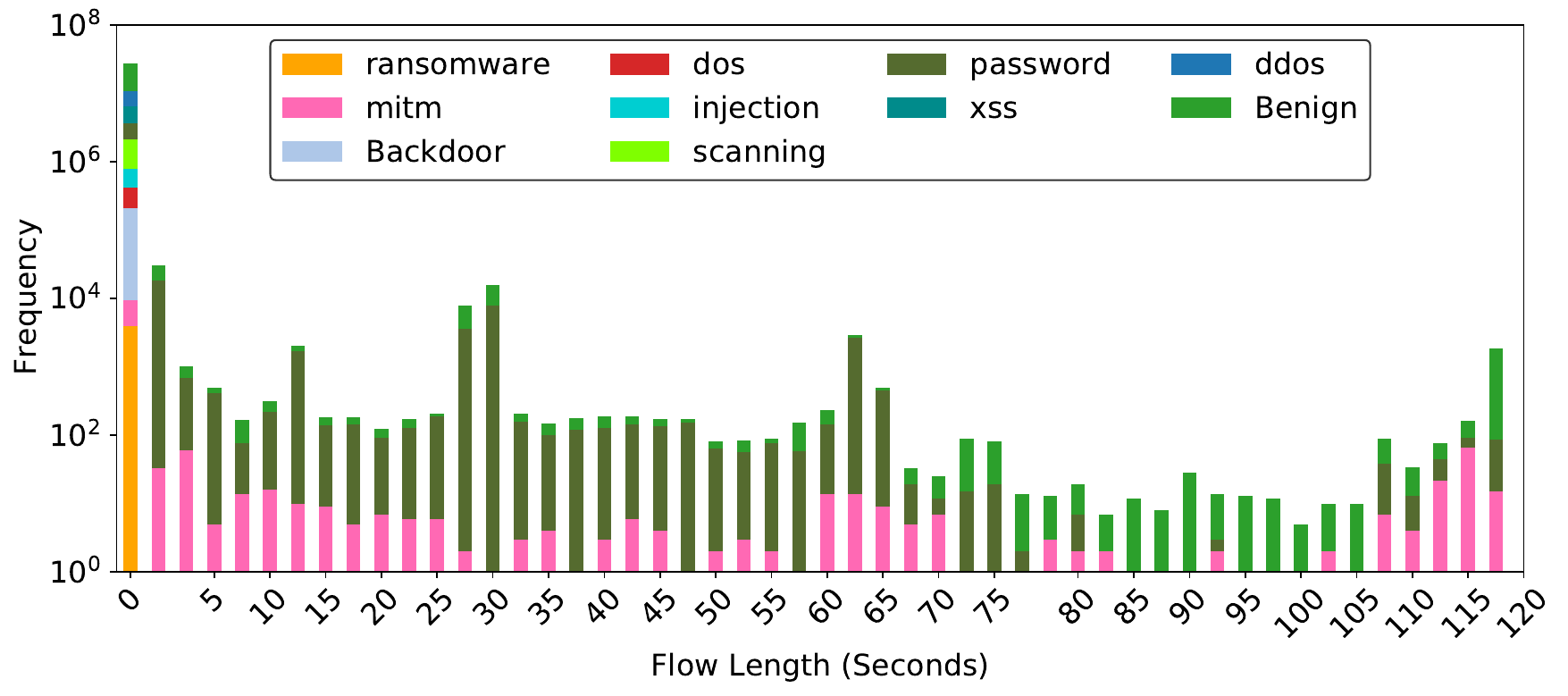}}\hspace{0.01cm}
    \subfloat[\centering][UNSW-NB15]
    {\includegraphics[width=0.49\linewidth]{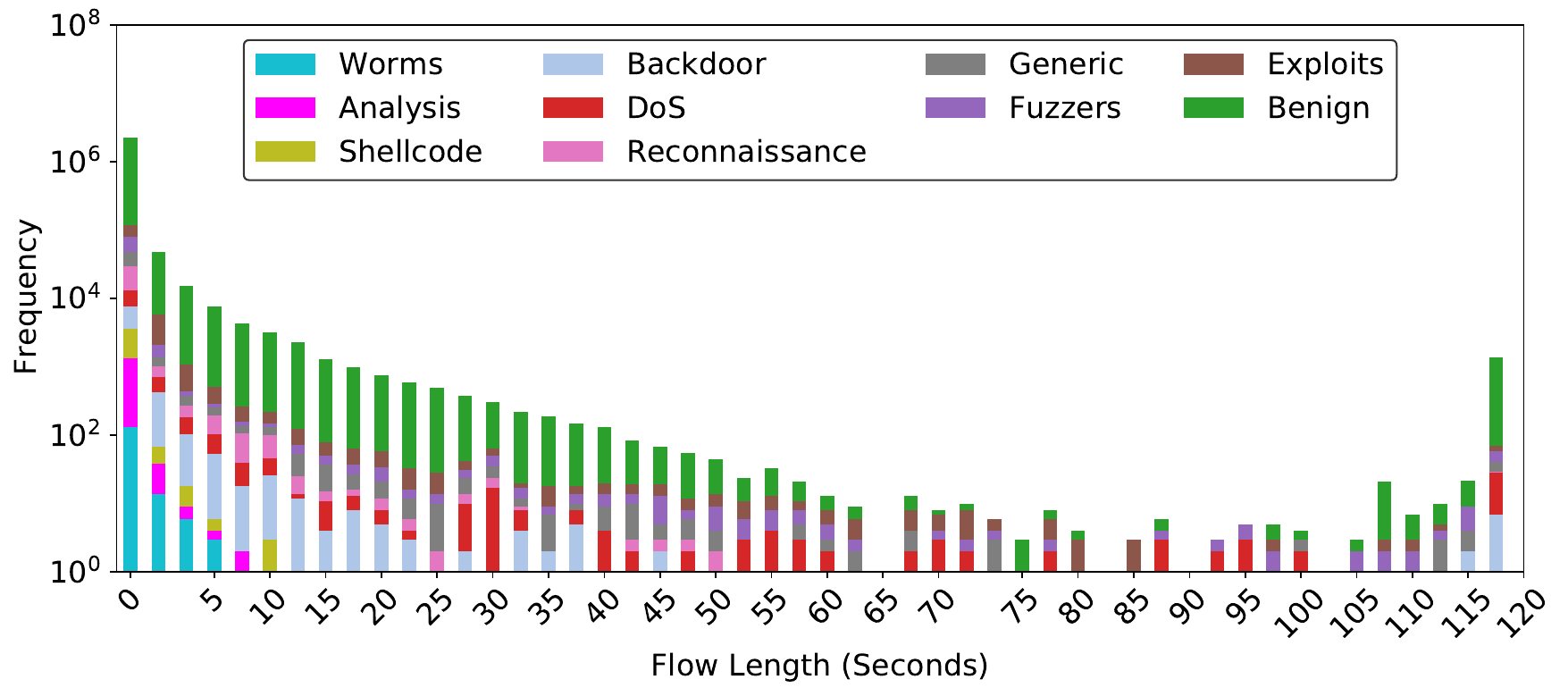}}
    
    \caption{Flow length distribution in NF3-Datasets. The x-axis represents the length of flows in seconds, while the y-axis represents the frequency of a length, i.e., the number of flows with the same flow length.}
    \label{FLD}
\end{figure}

\subsection{Flow Length Distribution}

The analysis of flow length distribution (FLD) across various datasets provides critical insights into the behaviour of network traffic under both benign and malicious conditions. This subsection visualises and discusses FLD for our NetFlow datasets.  In Figure~\ref{FLD}, each plot presents the frequency of flow lengths, aggregated into predefined bins (50 bins), across all the classes of traffic. However, the nProbe tool, by default, is configured to export flow data in intervals not exceeding two minutes. This is a standard configuration that allows for efficient flow data collection without overwhelming the system with excessive data~\citep{Ntop}. The 2-minute interval is chosen to provide a reasonable level of detail while minimizing system resource consumption.

\begin{figure}[!b]
    \centering
    \subfloat[\centering][BoT-IoT]
    {\includegraphics[width=0.49\linewidth]{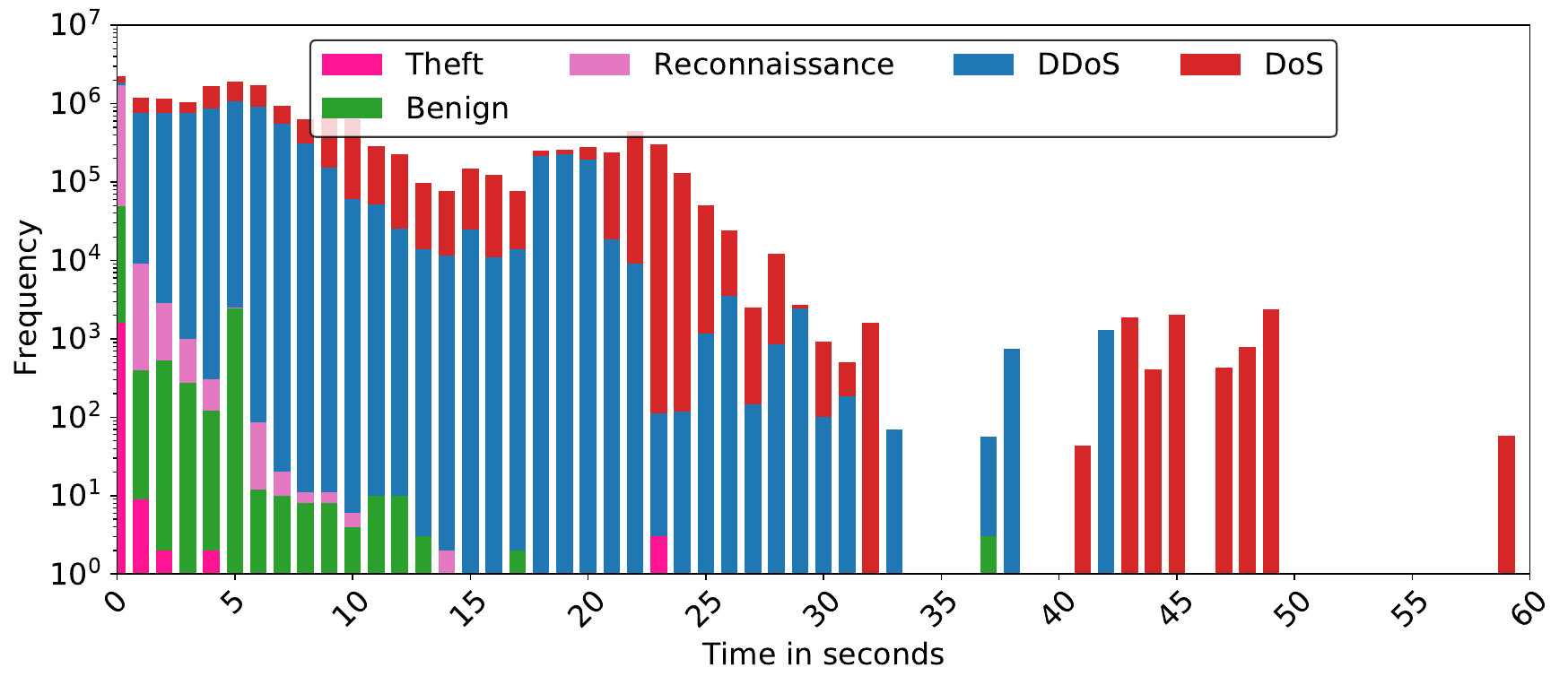}}\hspace{0.01cm}
    \subfloat[\centering][CSE\_CIC\_IDS2018]
    {\includegraphics[width=0.49\linewidth]{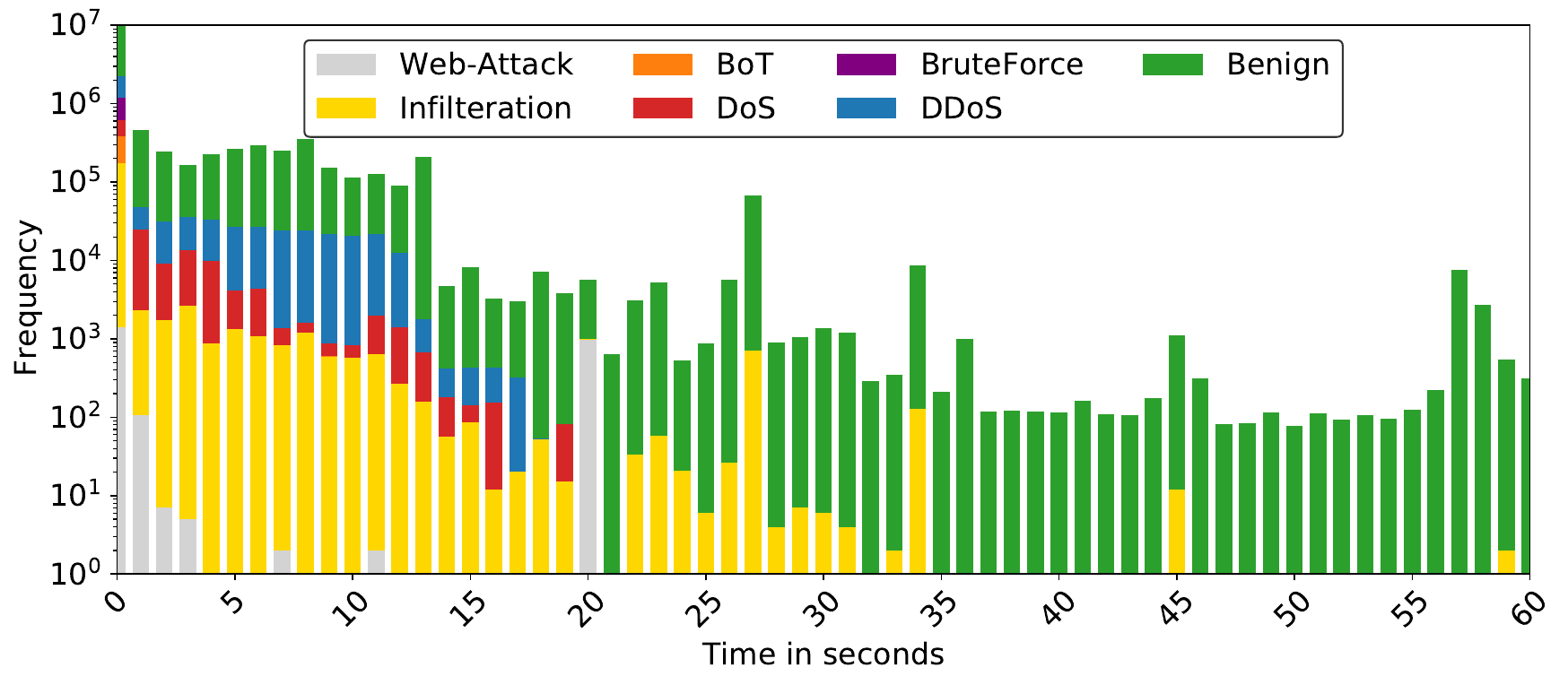}}\hspace{0.01cm}
    
    \subfloat[\centering][ToN-IoT]
    {\includegraphics[width=0.49\linewidth]{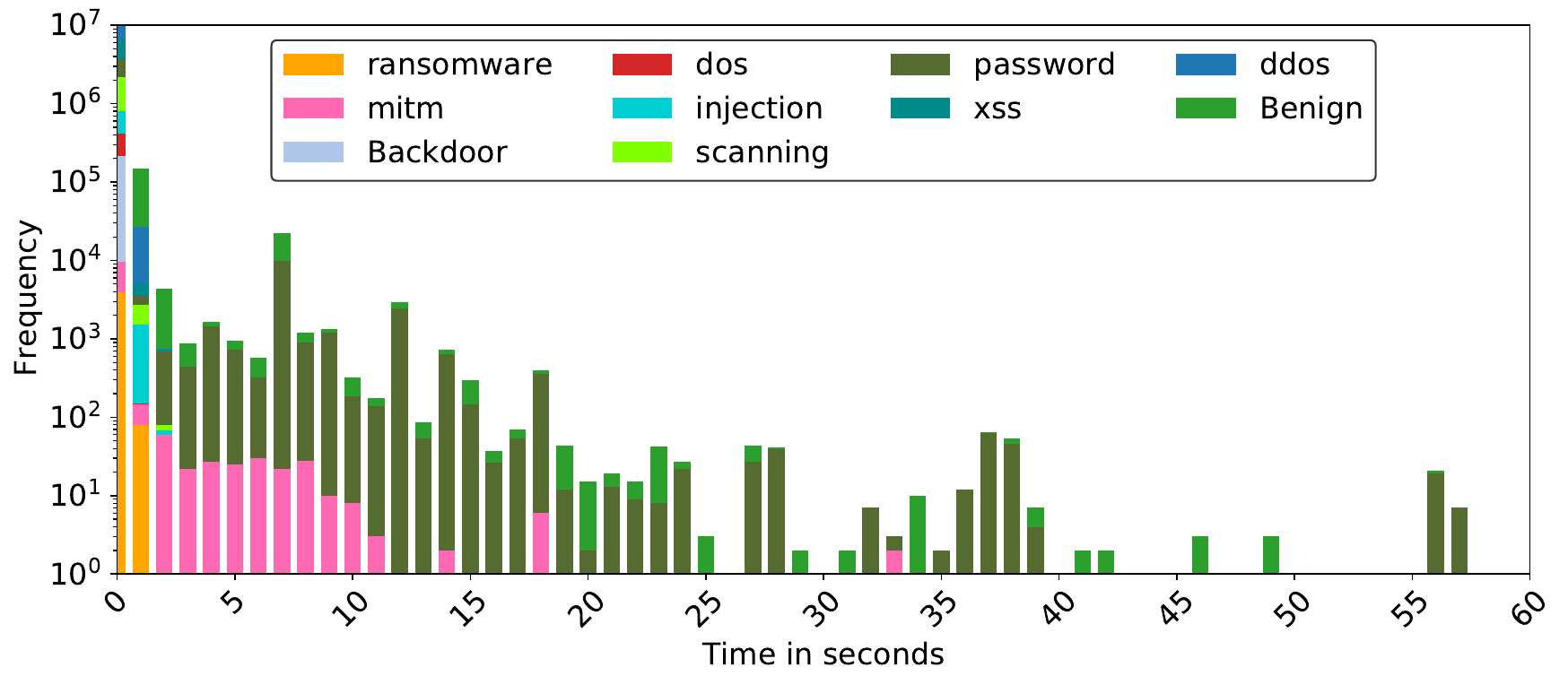}}\hspace{0.01cm}
    \subfloat[\centering][UNSW-NB15]
    {\includegraphics[width=0.49\linewidth]{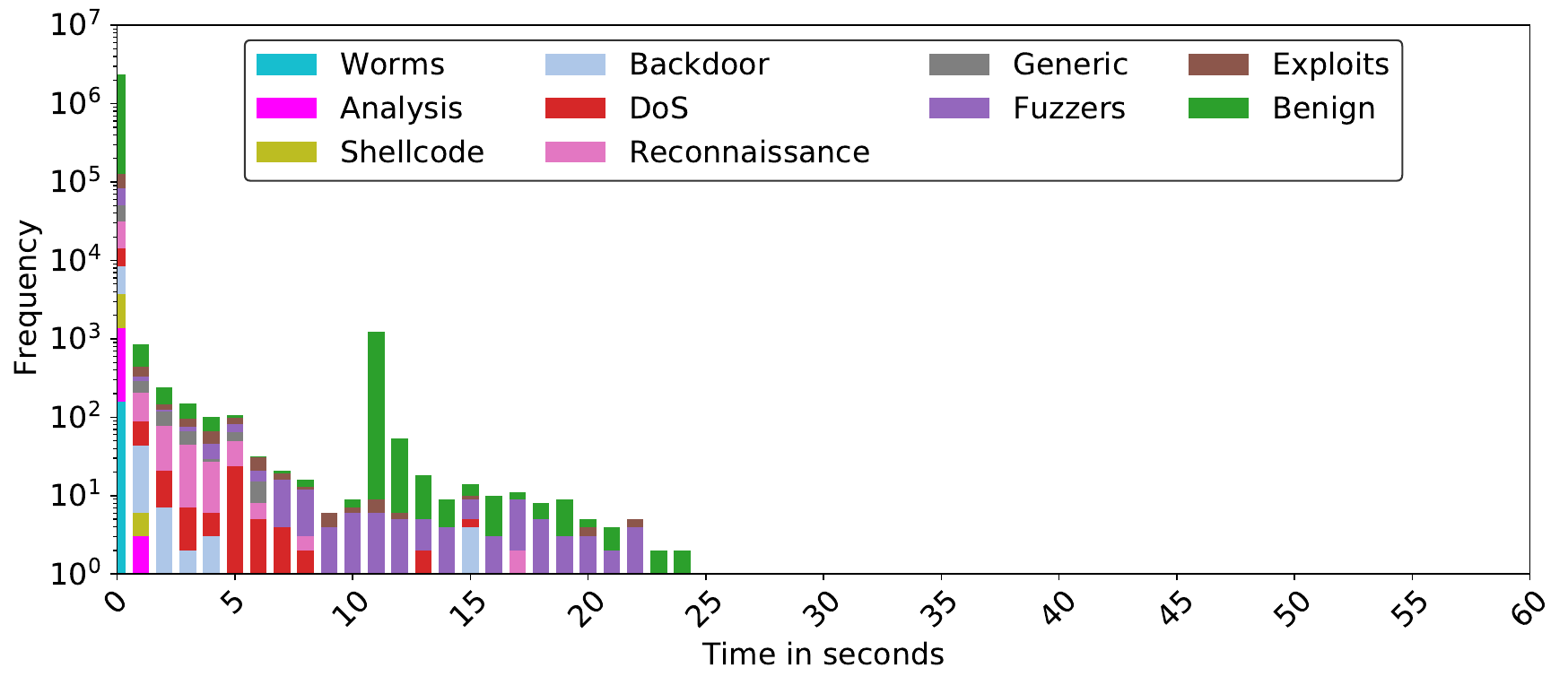}}
    
    \caption{Average distribution for Inter-Packet arrival time from source to destination.}
    \label{Inter-Packet_SRC}
\end{figure}

\begin{figure}[!t]
    \centering
    \subfloat[\centering][BoT-IoT]
    {\includegraphics[width=0.49\linewidth]{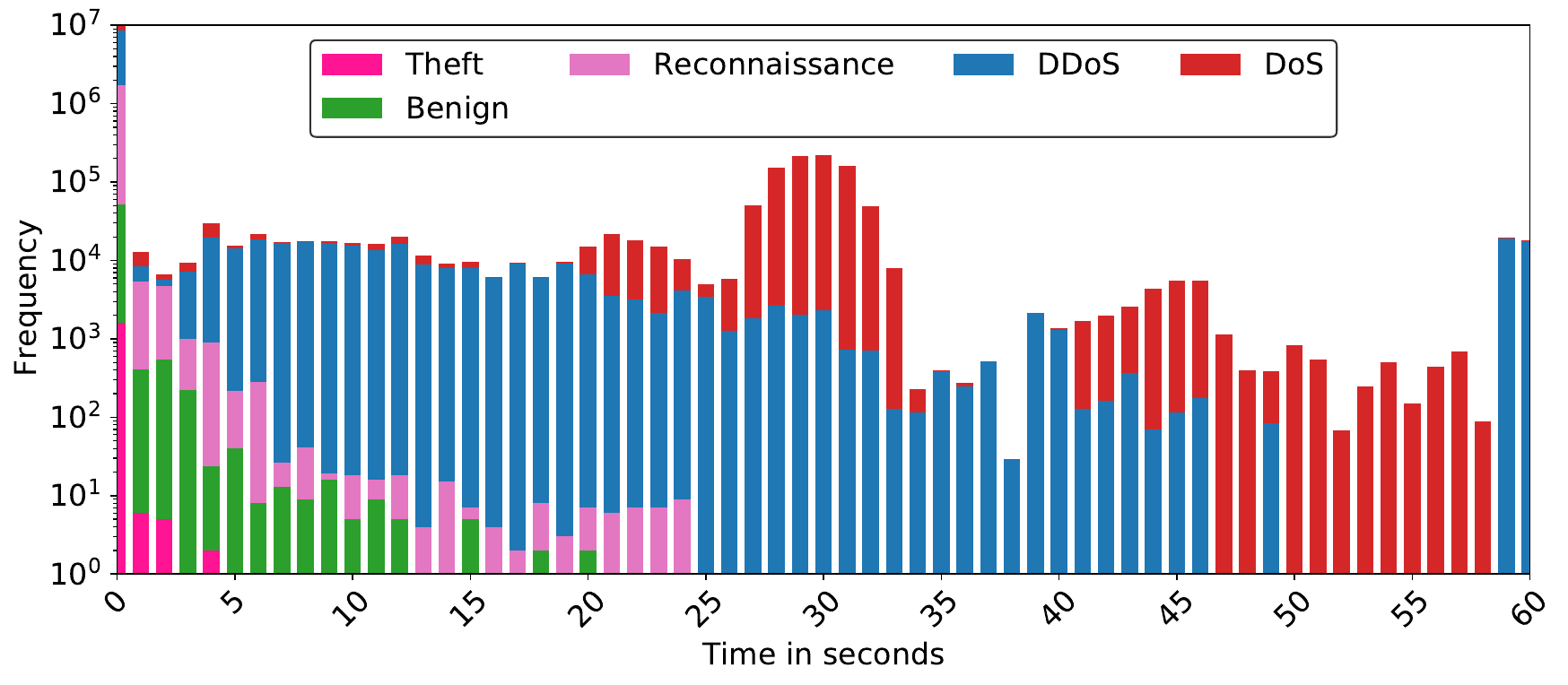}}\hspace{0.01cm}
    \subfloat[\centering][CSE\_CIC\_IDS2018]
    {\includegraphics[width=0.49\linewidth]{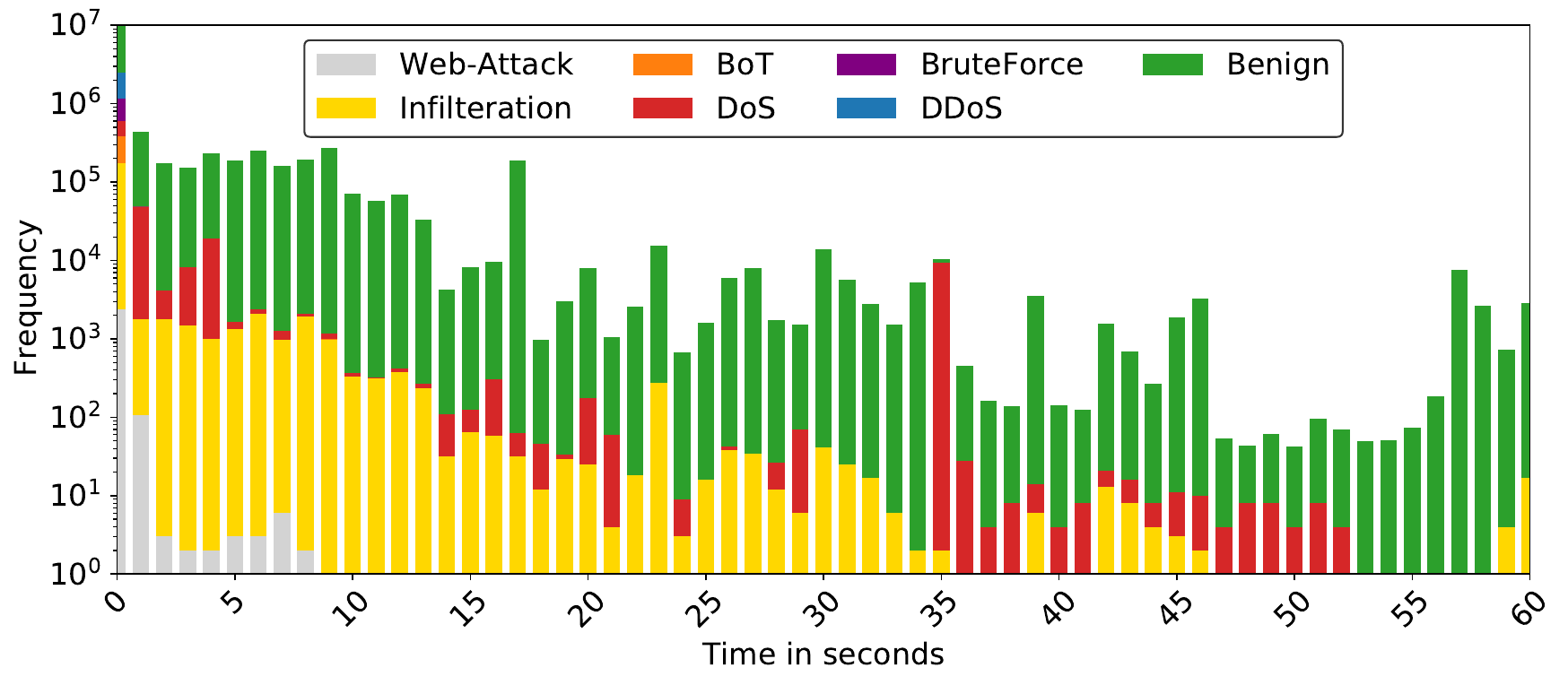}}\hspace{0.01cm}
    
    \subfloat[\centering][ToN-IoT]
    {\includegraphics[width=0.49\linewidth]{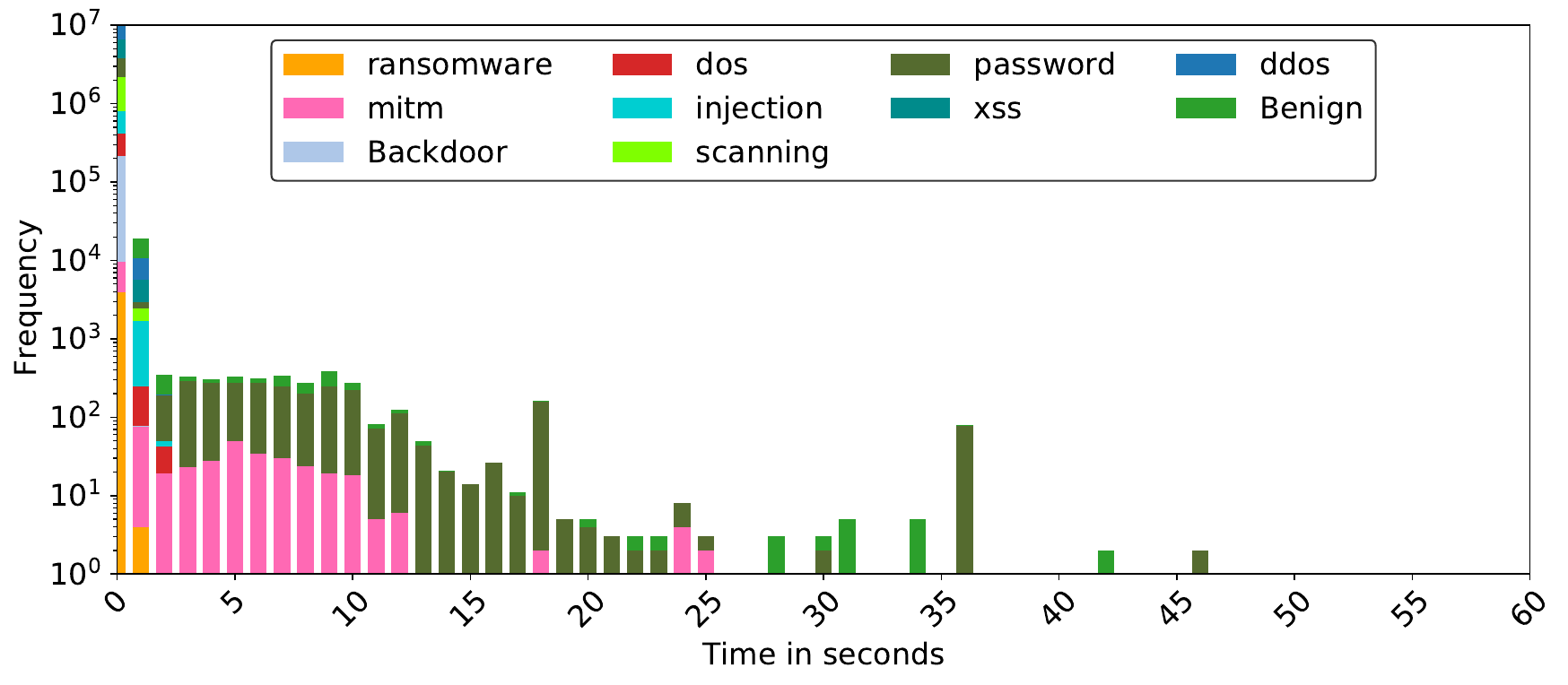}}\hspace{0.01cm}
    \subfloat[\centering][UNSW-NB15]
    {\includegraphics[width=0.49\linewidth]{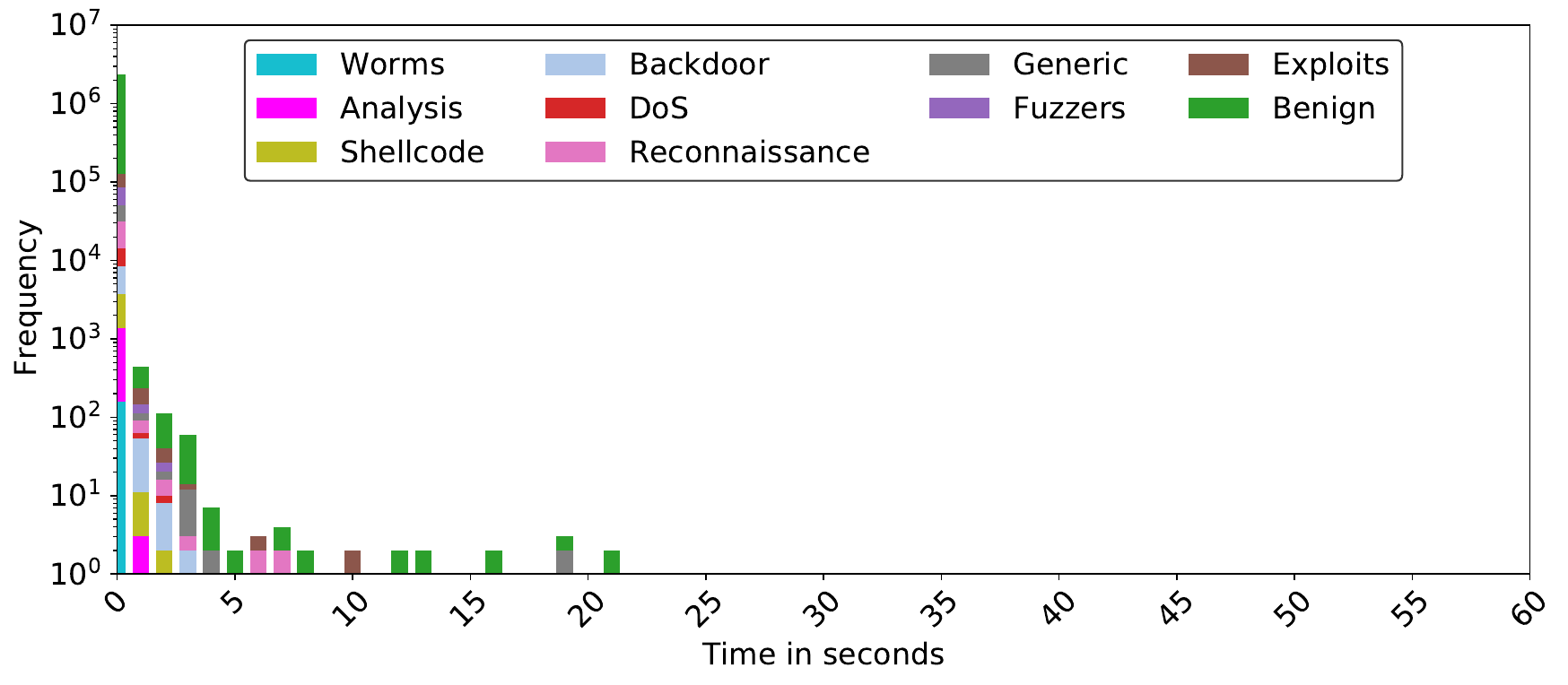}}
    
    \caption{Average distribution for Inter-Packet arrival time from destination to source.}
    \label{Inter-Packet_DST}
\end{figure}

In NF3-UNSW-NB15, benign flows predominantly appear in shorter-length bins, suggesting quick, routine communications typical in normal network operations. In contrast, attack flows such as Backdoor and Worms exhibit longer flow lengths, indicating sustained connections possibly used for data exfiltration or maintaining persistent threats within the network. Benign flows in NF3-BoT-IoT  are consistently short, reflecting typical user-generated traffic. However, DDoS and DoS attacks show a broad distribution across all flow lengths, highlighting their disruptive nature, which is characterised by both short and burst-like flows and prolonged attack durations to exhaust network resources. In the NF3-CSE-CIC-IDS2018 dataset, the flow lengths of benign traffic are moderately spread, indicating a variety of normal operations. Attack types such as DDoS and Brute Force attacks show significant occurrences at mid-range flow lengths, suggesting these attacks involve sequences of interactions that may be a part of the attack strategy to probe or compromise the network.
Lastly, FLD in NF3-ToN-IoT highlights notable distinctions between benign traffic and attack types such as MITM, Injection, and Password attacks. The majority of benign flows are short, which is consistent with normal operational traffic. Attack flows, particularly Password and MITM, demonstrate variability in their length distributions, reflecting the diverse tactics employed, from quick compromise attempts to more extended unauthorised access.

Across all datasets, the benign flows commonly populate the shortest flow length bins, reflecting typical, efficient network communications. Attack flows, depending on their nature, either mimic benign profiles or exhibit extended lengths, indicative of malicious activities. Such patterns are crucial for developing effective security measures, as they allow for the characterization of traffic based on flow length, enhancing anomaly detection capabilities.

\subsection{Inter-Packet Arrival time}

Analysing the histograms for the distribution of IAT provides valuable insights into how network behaviours are influenced by different types of network activities and attacks. Consistent IAT intervals typically indicate smooth traffic flow, while variability can reveal issues such as congestion or uneven data transmission. In this subsection, we specifically focus on the average IAT across the four NetFlow datasets. Figures~\ref{Inter-Packet_SRC} and~\ref{Inter-Packet_DST} display the distributions of these averages, illustrating the timing dynamics across all communications between sources and destinations within each dataset. Figure~\ref{Inter-Packet_SRC} shows IAT distribution from source to destination across the four datasets and similarly, Figure~\ref{Inter-Packet_DST} shows the opposite direction from destination to source. These plots highlight the variability in IAT across benign and malicious traffic, offering clues into network dynamics under various conditions.

Each dataset reveals unique IAT patterns for different attack types. For example, the ToN-IoT dataset shows distinct peaks for more sophisticated attacks like MITM (Man-in-the-Middle) and Backdoor at specific IAT intervals, possibly reflecting the tactical nature of these attacks, which may involve periodic signalling or data exfiltration activities. Similarly, the UNSW-NB15 dataset demonstrates how diverse attack types like Worms, Shellcode, and Exploits are distributed across various IAT ranges, highlighting the varied timing strategies used in different exploits. In NF3-BoT-IoT, the benign traffic is characterised by shorter IATs, frequently occurring at lower millisecond ranges, which is indicative of regular, uninterrupted network flow. In contrast, malicious activities such as DoS and DDoS attacks show a wider distribution of average IAT values, with notable peaks at higher intervals, reflecting the irregular timing patterns typical of such attacks that disrupt normal network traffic patterns.

Comparing these plots across datasets enriches our understanding of how different network environments or attack vectors can influence IAT distributions. It also underscores the importance of considering context and environment when analysing network traffic, as the same type of attack may exhibit different IAT characteristics in different datasets.


\subsection{Number of Flows vs. Time}
\begin{figure}[!b]
    \centering
    \subfloat[\centering][BoT-IoT Day 1]
    {\includegraphics[width=0.49\linewidth]{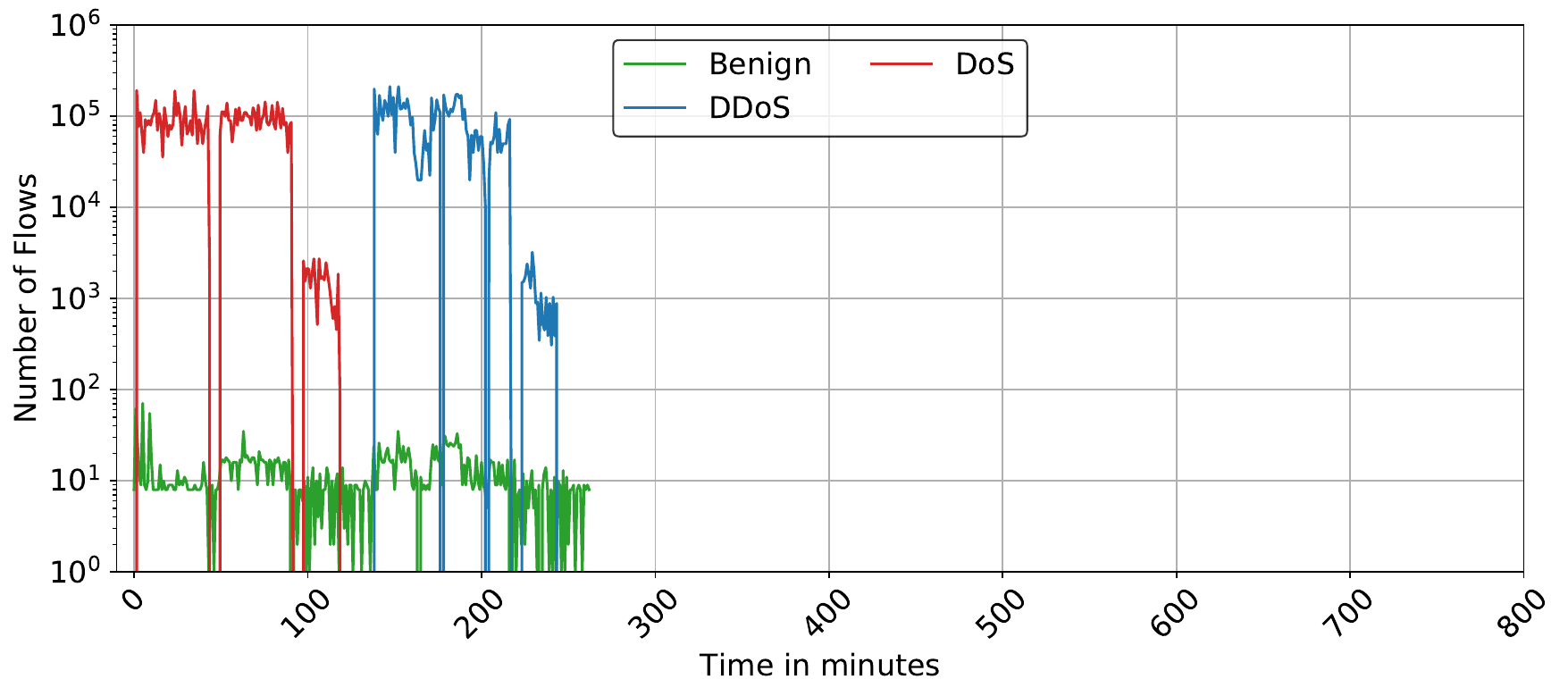}}\hspace{0.01cm}
    \subfloat[\centering][CSE\_CIC\_IDS2018 Day 5]
    {\includegraphics[width=0.49\linewidth]{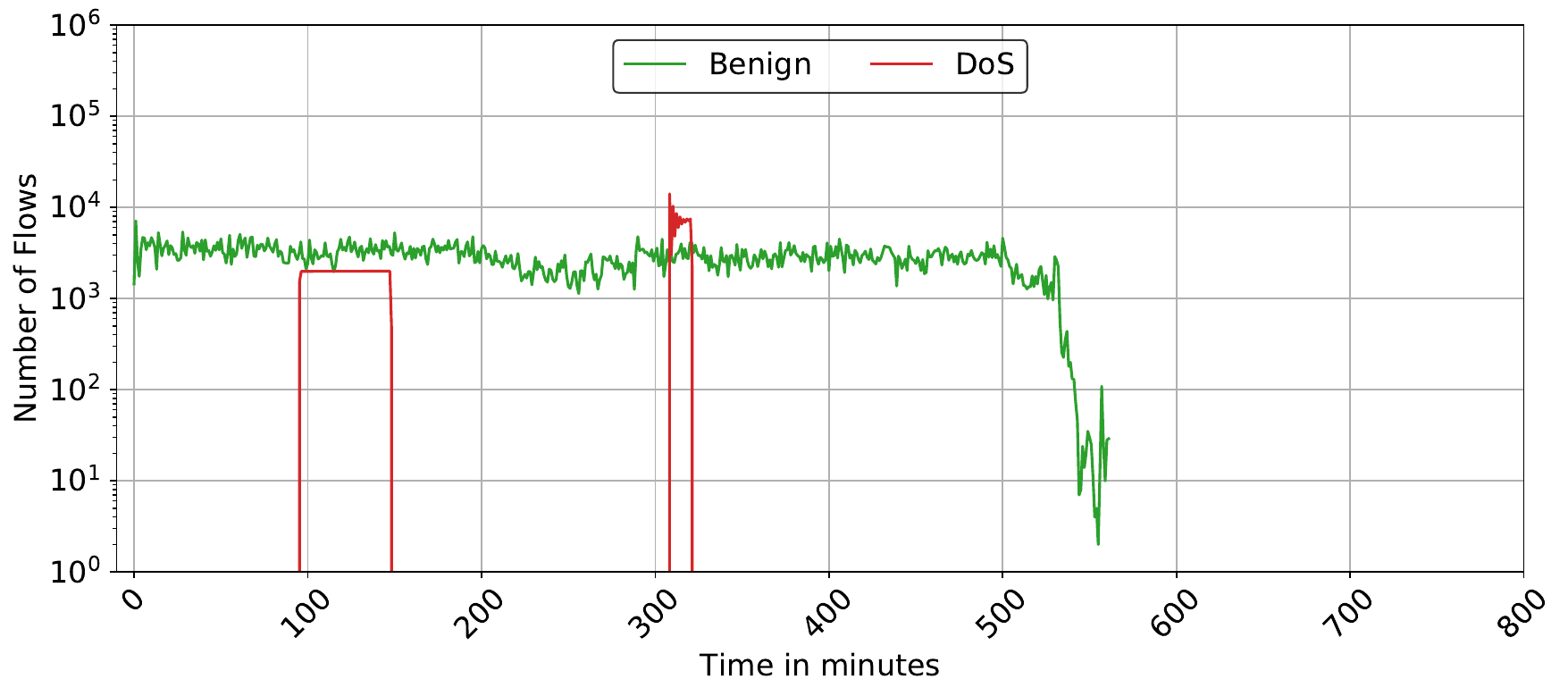}}\hspace{0.01cm}
    
    \subfloat[\centering][ToN-IoT Day 5]
    {\includegraphics[width=0.49\linewidth]{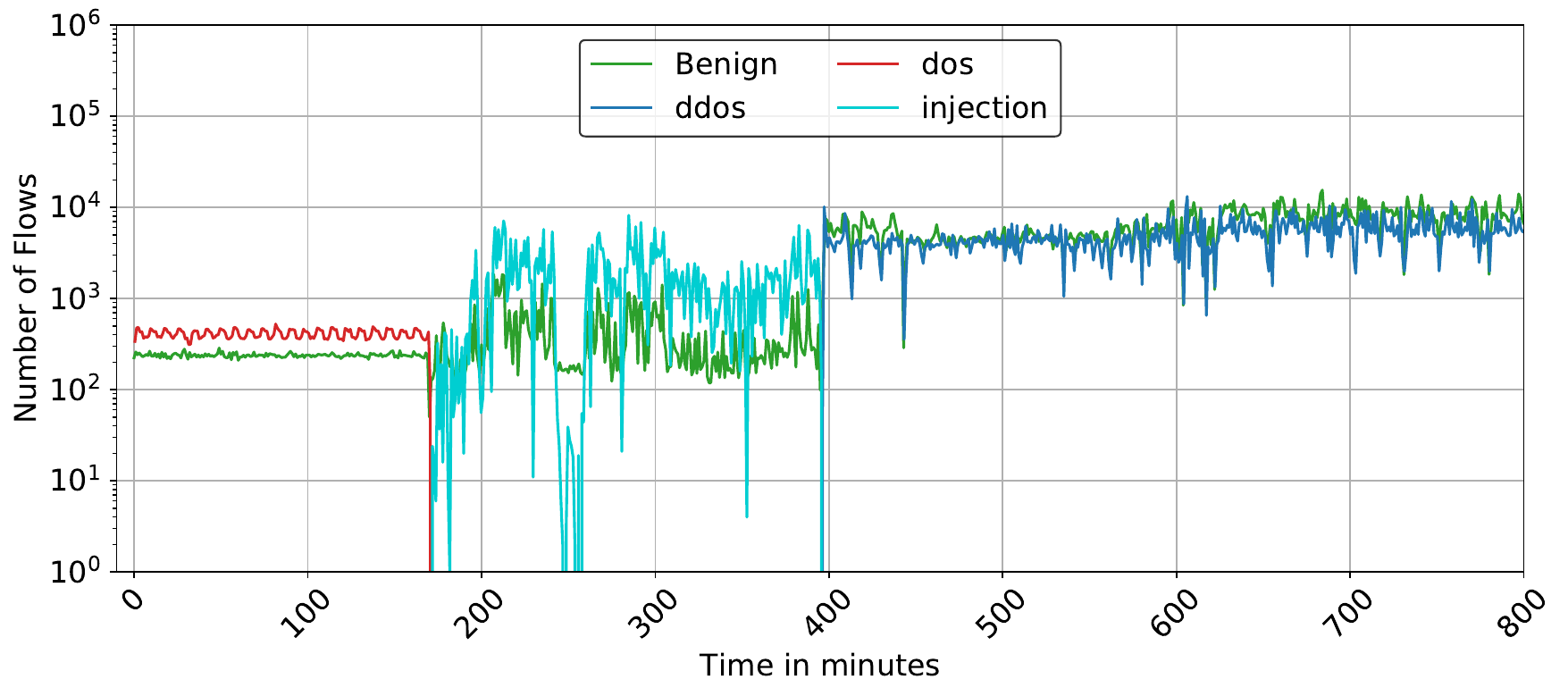}}\hspace{0.01cm}
    \subfloat[\centering][UNSW-NB15 Day 1]
    {\includegraphics[width=0.49\linewidth]{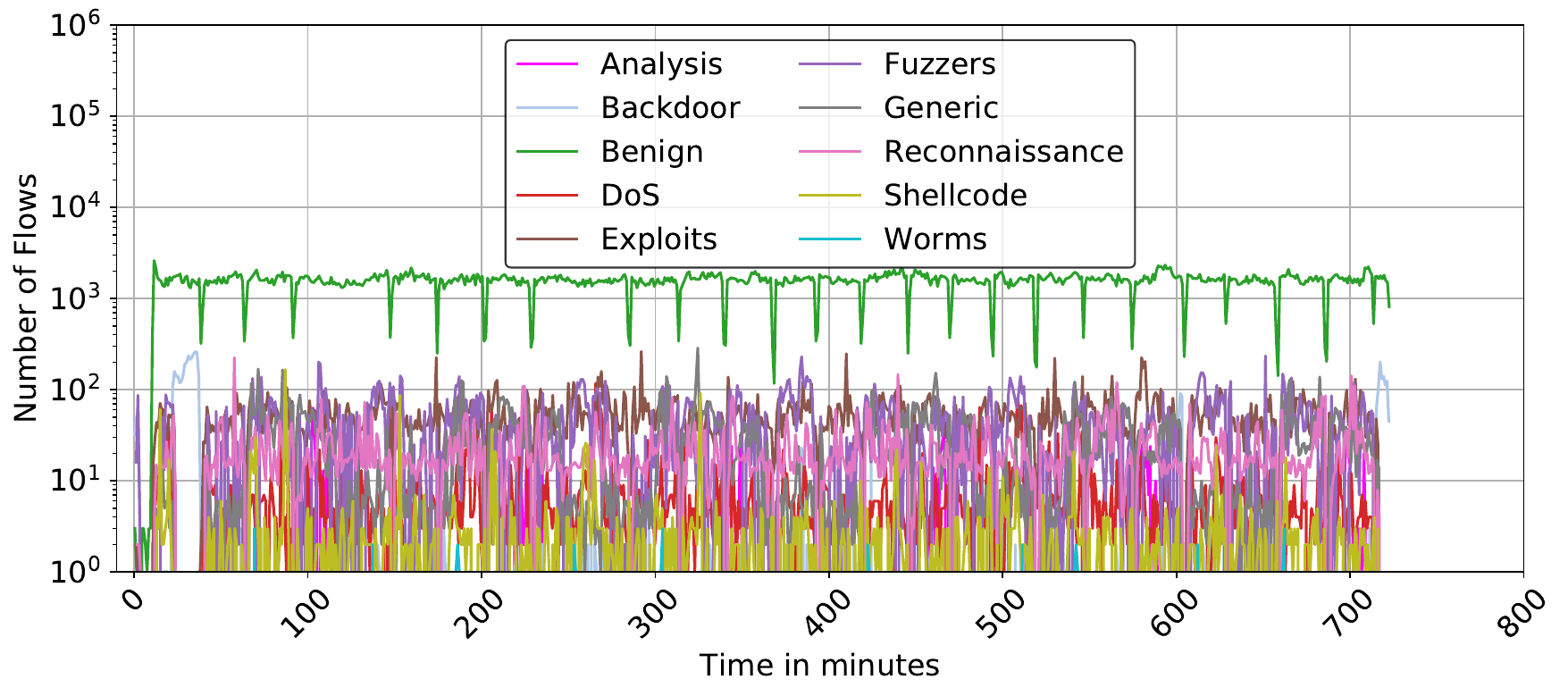}}
    
    \caption{Temporal Distribution of Network Traffic Across Four Datasets. This figure illustrates the minute-by-minute network traffic flow for NF3-Datasets on representative days, showcasing the onset, duration, and termination of various attack classes alongside benign traffic.}
    \label{Trafficc_Distribution}
\end{figure}

When analysing traffic over time, it is important to track the distribution of attack classes within the relevant time intervals. This helps in understanding how many flows are labelled as benign or malicious, providing a clearer picture of the traffic behaviour. In this subsection, we represent the traffic as a time series for each attack class to pinpoint their exact occurrence times. Typically, most datasets were recorded over multiple days to simulate real-world conditions. As depicted in Figure~\ref{Trafficc_Distribution}, we chose one representative day from each dataset, aggregating the traffic data per minute and displaying the volume on a logarithmic scale to enhance the clarity of visual interpretation.

Starting with day 1 of NF3-UNSW-NB15, all attack classes occur concurrently throughout the day, providing a complex overlay of multiple threats, which is characteristic of sophisticated real-world attack scenarios. This simultaneous occurrence requires further analysis techniques to isolate and identify individual attack vectors. Another observation from NF3-BoT-IoT Day 1 is the clear periods of intense DDoS and DoS attacks, with sharp increases in flow counts, followed by periods of lower activity. This pattern suggests the attacks were launched in waves, a common tactic in denial-of-service attacks to overwhelm systems periodically. On the fifth day of the NF3-CSE-CIC-IDS2018 dataset, the distribution reveals a dominant presence of benign traffic, with intermittent spikes in DoS attack flows. The attack patterns appear as short-lived bursts rather than continuous flooding, suggesting controlled execution, possibly mimicking real-world attack scenarios or stress-testing conditions. Lastly, NF3-ToN-IoT on day 5 displays separate and distinct instances of DDoS, DoS, and Injection attacks along with periods of benign activity. Throughout the day, benign traffic remains consistent and predominantly at a lower flow level, which is typical of a synthetic dataset designed to maintain a baseline for comparison. This distribution suggests that while attacks are not related or overlapping, the dataset effectively captures distinct and varied attack dynamics within the same day, allowing for the analysis of each threat type under controlled conditions.

While the analysis presented focuses on a single representative day for each dataset, similar examinations were conducted across all active days within each dataset. This comprehensive analysis is crucial for developing a robust understanding of the variability and consistency of network attack behaviours over extended periods. The results underscore the diversity in attack methodologies and their temporal characteristics, which can vary not just from day to day but also from one dataset to another.

After representing the whole period of each dataset, we found that most attack classes were implemented separately on different days. However, an exception is observed in the NF3-UNSW-NB15 dataset, where all attacks were injected simultaneously. While having multiple attacks simultaneously can occur in real-life scenarios, it is recommended for researchers to analyse each class individually to better understand its pattern. Table~\ref{daily_information} catalogues, in detail, the number of active days for each dataset along with the specific attacks implemented on those days. This tabulation aids in quantifying the extent and variety of network attacks captured in the datasets, providing a foundational reference for further analysis or model training.

\begin{figure}[!b]
    \centering
    \subfloat[\centering][BoT-IoT Day 1]
    {\includegraphics[width=0.49\linewidth]{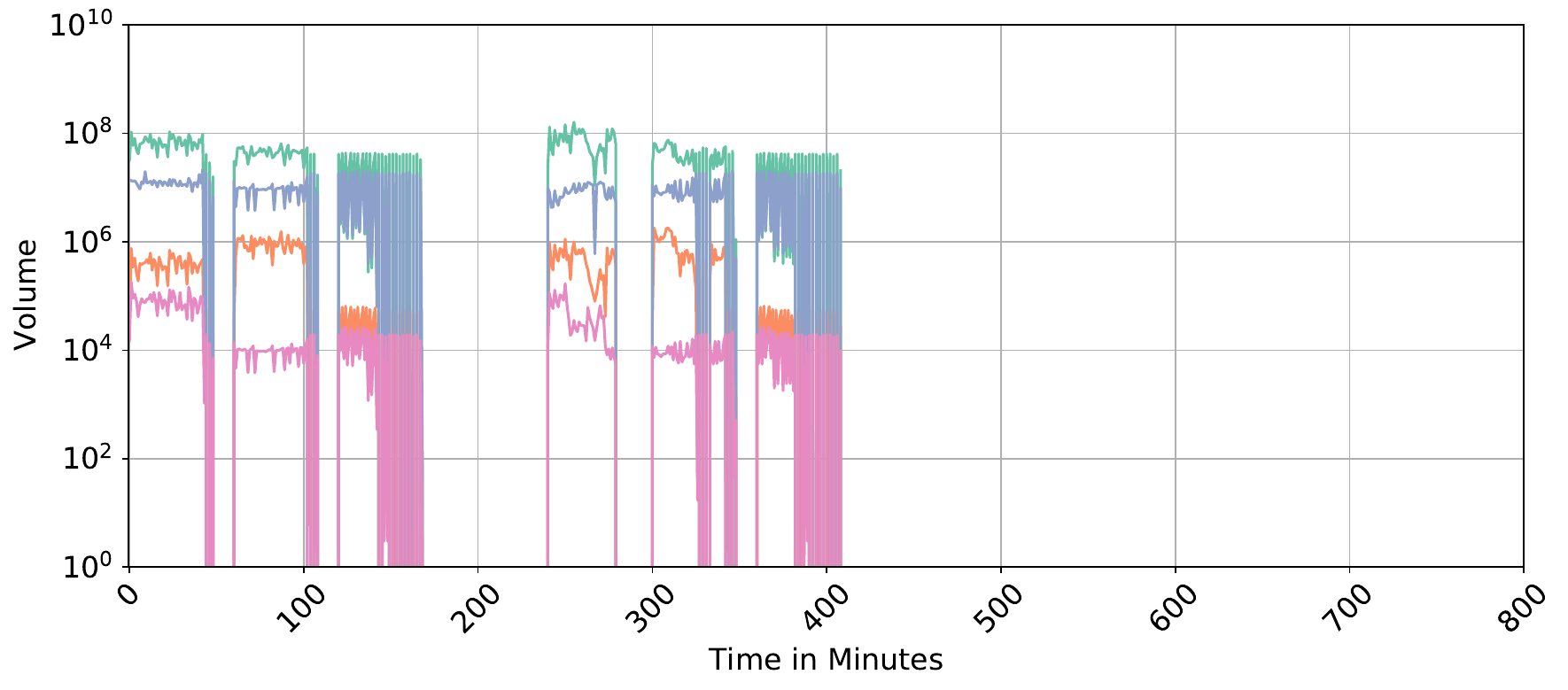}}\hspace{0.01cm}
    \subfloat[\centering][CSE\_CIC\_IDS2018 Day 5]
    {\includegraphics[width=0.49\linewidth]{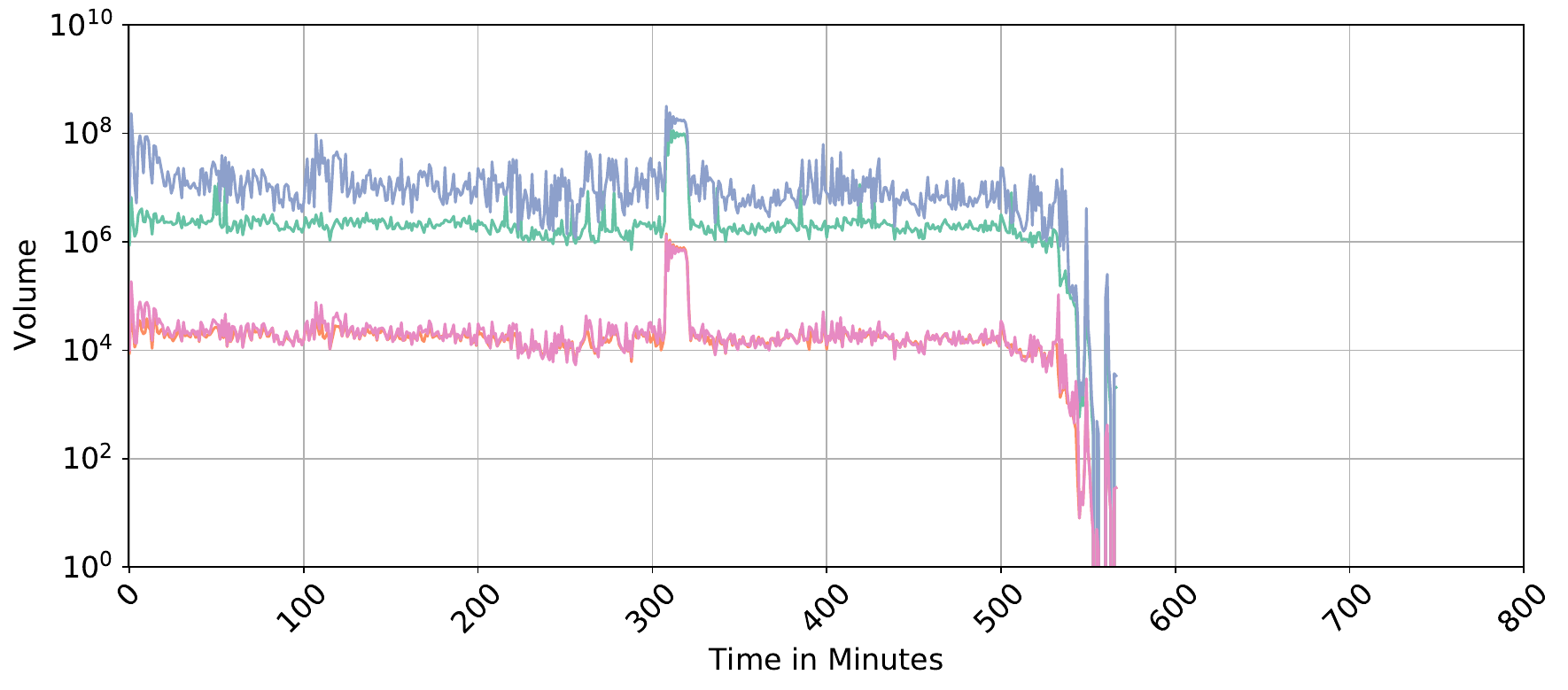}}\hspace{0.01cm}
    
    \subfloat[\centering][ToN-IoT Day 5]
    {\includegraphics[width=0.49\linewidth]{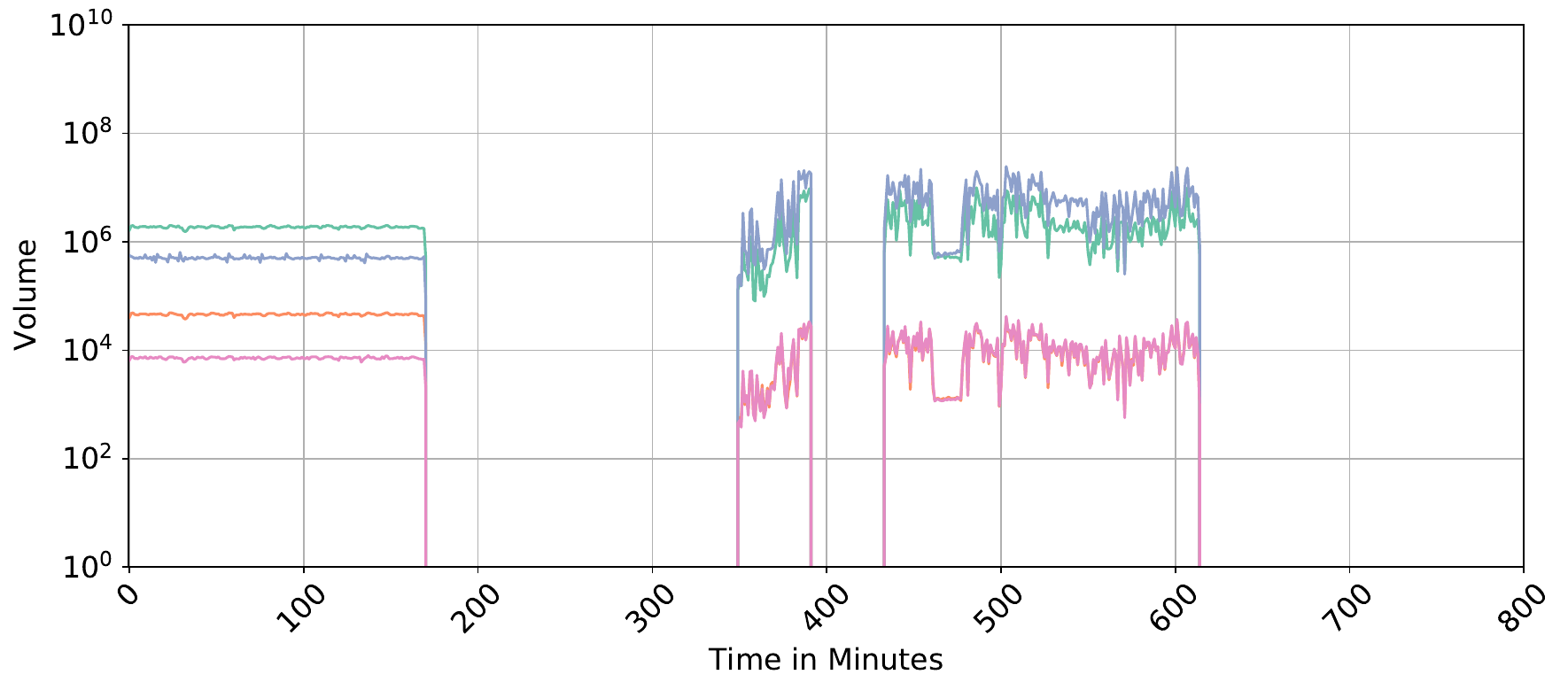}}\hspace{0.01cm}
    \subfloat[\centering][UNSW-NB15 Day 1]
    {\includegraphics[width=0.49\linewidth]{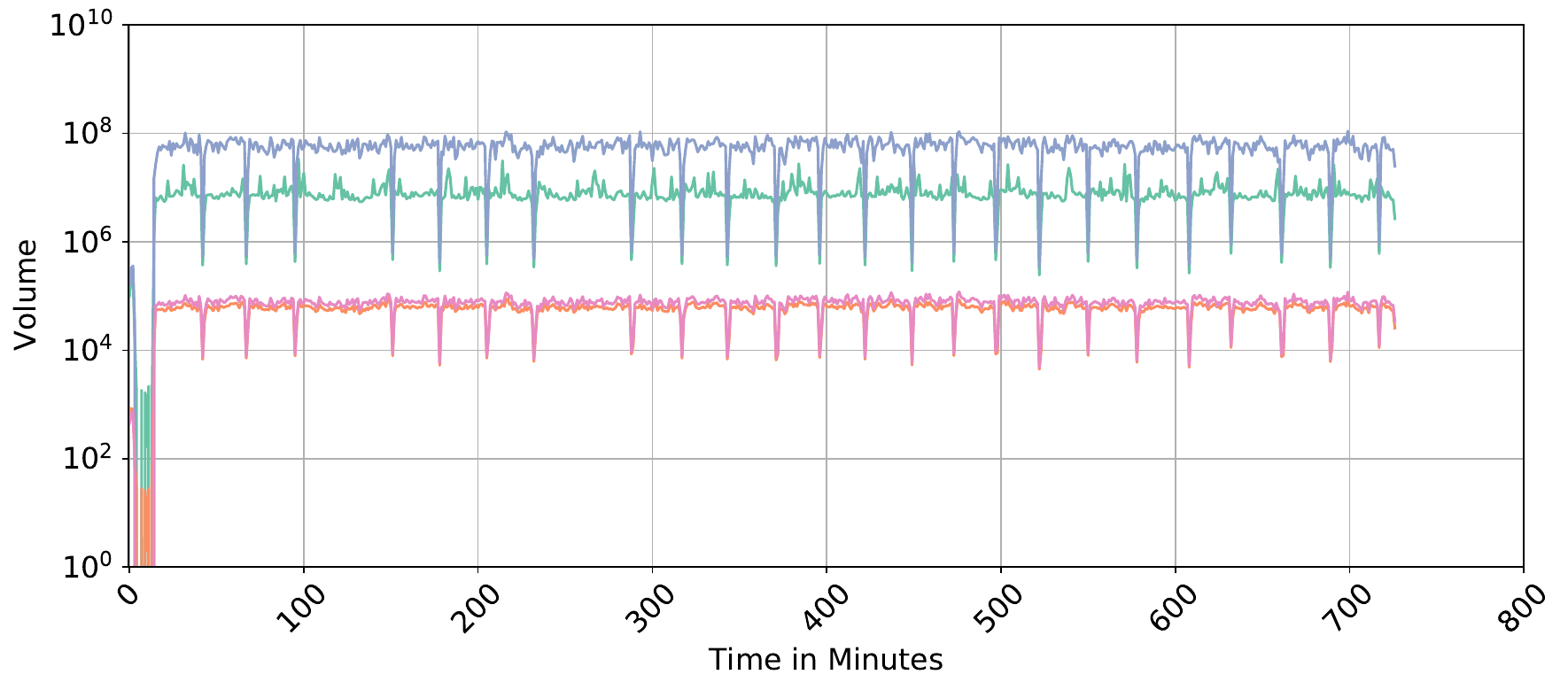}}

    \vspace{0.02cm} 
    \includegraphics[width=0.6\linewidth,trim=10 50 10 50,clip]{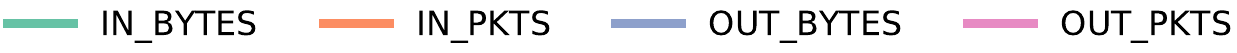}

    \caption{Time series representation of numerical fields in NF3-Datasets: IB, OB, IP, and OP. The x-axis represents time aggregated in minutes, while the y-axis shows the volume of each feature, illustrating fluctuations and patterns in network traffic over time.}
    \label{Numerical_TS_Figures}
\end{figure}

\begin{table*}[!t]
    \centering
    \scriptsize
\caption{Attacks Implemented on Active Days for Each Dataset}    \label{daily_information}
    \begin{tabular}{lcccc}
        \toprule
        \textbf{Days} & \textbf{NF3-UNSW-NB15} & \textbf{NF3-CSE-CIC-IDS2018} & \textbf{NF3-ToN-IoT} & \textbf{NF3-BoT-IoT}\\
        \midrule
         1 & All & BruteForce & Benign-Only & Reconnaissance\\
        2 & All & DoS & Benign-Only & Reconnaissance\\
        3 &  Benign-Only  & DoS & Benign-Only & Reconnaissance\\
        4 &  —  & DDoS & Scanning & DoS, DDoS\\
        5 &  — & DDoS & DoS, Scanning & Theft\\
        6 &  —  & Web-Attack & DDoS, Injection, DoS & Theft \\
        7 &  — & Web-Attack & DDoS, Password &  — \\
        8 &  — &  Benign-Only & XSS, Password& — \\
        9 &  — & Infiltration & Backdoor, Ransomware& — \\
        10 &  —  &  Infiltration & MITM, Backdoor& — \\
        11 &  —  & BoT &  — & — \\
        \bottomrule
    \end{tabular}
\end{table*}



\subsection{Time-series Representation of NetFlow Features}
Monitoring network traffic volume over time is essential for understanding network behaviour and identifying trends or irregularities that may not be apparent in static analysis. By analysing traffic as a time series, we can detect variations in network load, identify peak usage time intervals, and observe patterns of data flow across different time intervals. This continuous observation allows for a deeper understanding of normal traffic behaviour and helps to highlight anomalies or unusual patterns that could indicate underlying issues. In this subsection, we represent different numerical and categorical features from the datasets as time series to gain insights into the temporal dynamics of the traffic. This visualisation not only helps in understanding how these features distribute over time but also showcases the enhanced analysis capabilities introduced by adding temporal information into this version of the datasets.

\subsubsection{Numerical Fields}
In this analysis, we focus on four pivotal numerical features: IN\_BYTES (IB), IN\_PKTS (IP), OUT\_BYTES (OB), and OUT\_PKTS (OP). These features are instrumental in gauging the volume and flow of data moving into and out of the network, critical for deciphering overall traffic patterns~\citep{elephant1,elephant2,elephant3}. IB and OB measure the amount of data received and sent, respectively, offering insights into data load, bandwidth usage, and potential congestion points. Simultaneously, IP and OP count the number of packets transmitted, which is essential for assessing the efficiency of packet transmission, pinpointing any packet loss, and evaluating the balance of traffic flow.

To enable a thorough monitoring of network traffic over time, we aggregate these features by minute. This temporal granularity unveils detailed patterns and fluctuations in traffic that illuminate the network's performance and utilisation. For consistent and focused analysis, we have chosen the same single-day snapshots as in the previous section, as shown in Figure~\ref{Numerical_TS_Figures}.

The analysis of these time series reveals a symmetrical pattern between IB and OB, as well as between IP and OP, indicative of a balanced communication pattern within the network where the volume of incoming bytes and packets closely mirrors that of outgoing bytes and packets over time. This symmetry reflects a stable network environment where data inflow and outflow are consistent, suggesting effective network management and robust infrastructure.

Specific observations from the representative days across various datasets illustrate the nuanced dynamics of network traffic: NF3-ToN-IoT and NF3-CSE-CIC-IDS2018, both on Day 5, show consistent levels of IB and OB with sporadic spikes possibly linked to operational anomalies or specific events. In contrast, NF3-UNSW-NB15 Day 1 features a notable early spike in OB, suggesting an event like data exfiltration or a substantial data transfer, is potentially benign. Meanwhile, NF3-BoT-IoT Day 1 exhibits significant variability in OP, indicative of intermittent network attacks or disruptions, underscoring the susceptibility to external threats.

\subsubsection{Categorical Fields}
Categorical features, such as Origin/Destination IPs and Ports, offer valuable insights into the structure and behaviour of network traffic. By tracking the number of unique IPs and ports over time, we can better understand communication patterns, identifying which devices are actively engaged in the network. This also reveals the diversity of traffic whether it's distributed across many endpoints or concentrated on specific services. Additionally, monitoring these features helps detect unusual behaviour such as sudden increases in unique IPs or port activity which could indicate irregular network events~\citep{mining_Anomalies2005}. NIDS datasets often vary significantly in the number of unique IP addresses and ports they capture, reflecting differences in the scope and diversity of network traffic. The number of unique IPs and ports present in each of the proposed datasets is shown in Table~\ref{Categorical_Feilds_count}. 
\begin{table}[!b]
  \centering
  \scriptsize
  \caption{Count of unique categorical fields in NF3-Datasets}
  \label{Categorical_Feilds_count}
  \begin{tabular}{lcccc}
    \toprule
    \textbf{Dataset} & \textbf{Source IPs} & \textbf{Destination IPs} & \textbf{Source Ports} & \textbf{Destination Ports} \\
    \midrule
    NF3-UNSW-NB15 & 40 & 40 & 64,620 & 64,631 \\
    NF3-CSECIC-IDS2018 & 183,806 & 29,226 & 65,325 & 63,353 \\
    NF3-ToN-IoT & 15,396 & 9,011 & 65,536 & 65,536 \\
    NF3-BoT-IoT & 20 & 291 & 65,536 & 65,536 \\
    \bottomrule
  \end{tabular}
\end{table}

\begin{figure}[!t]
    \centering
    \subfloat[\centering][BoT-IoT Day 1]
    {\includegraphics[width=0.49\linewidth]{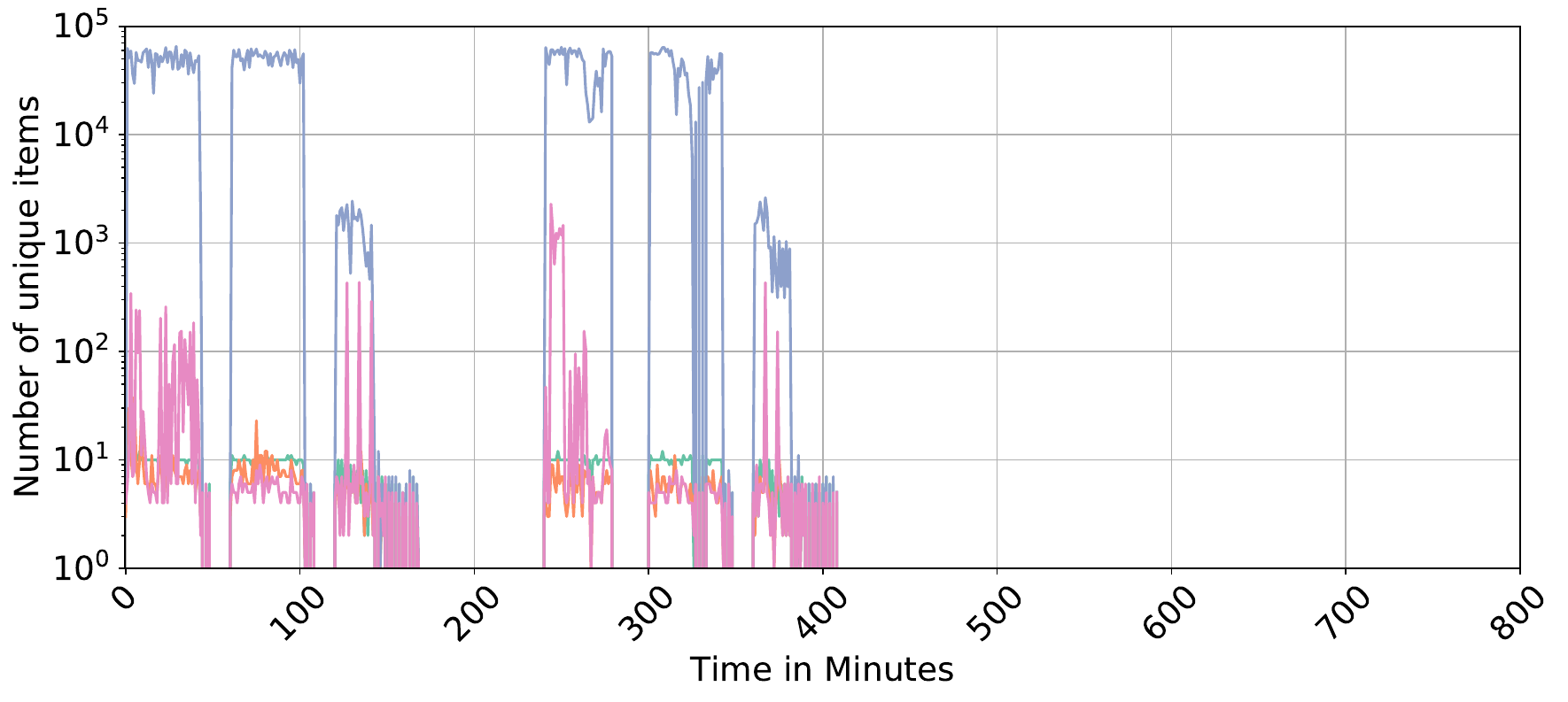}}\hspace{0.01cm}
    \subfloat[\centering][CSE\_CIC\_IDS2018 Day 5]
    {\includegraphics[width=0.49\linewidth]{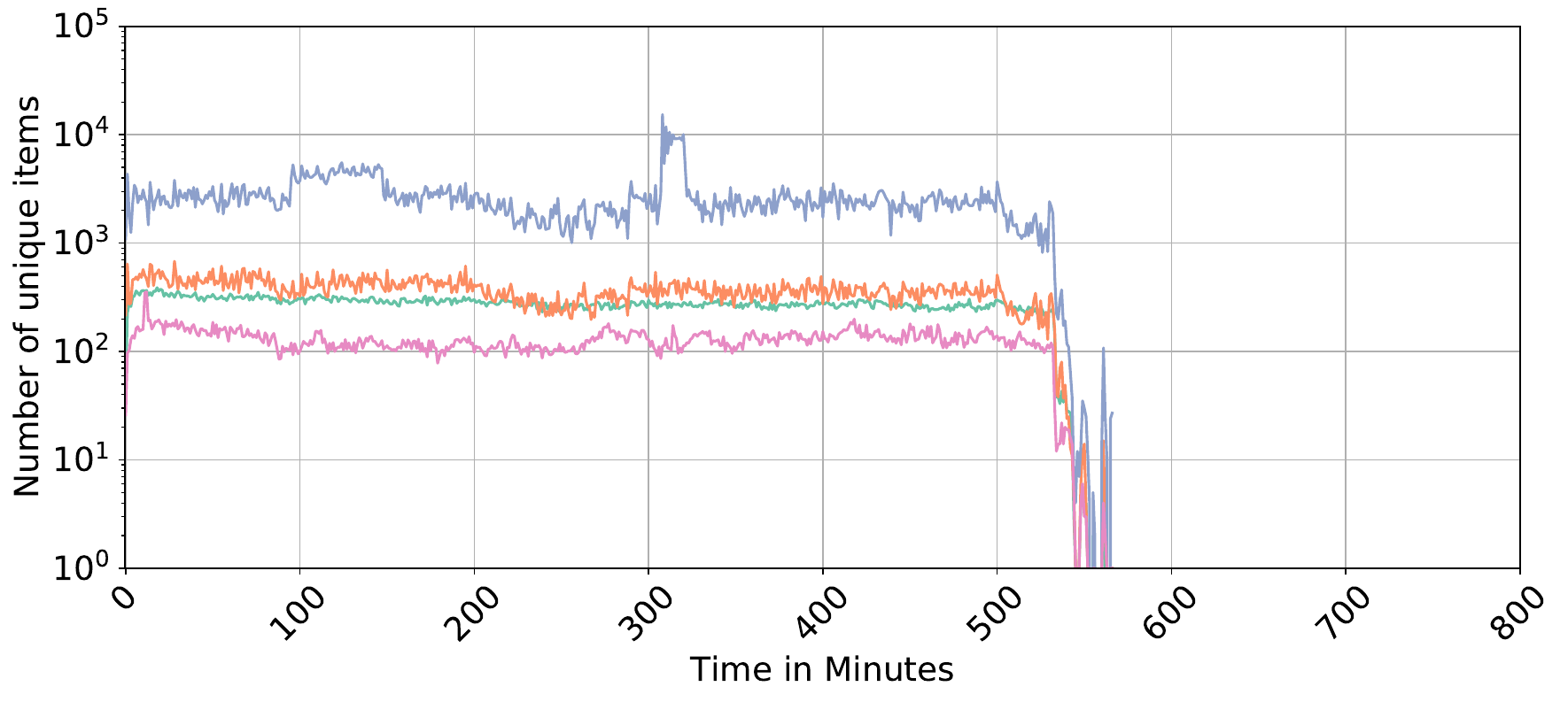}}\hspace{0.01cm}
    
    \subfloat[\centering][ToN-IoT Day 5]
    {\includegraphics[width=0.49\linewidth]{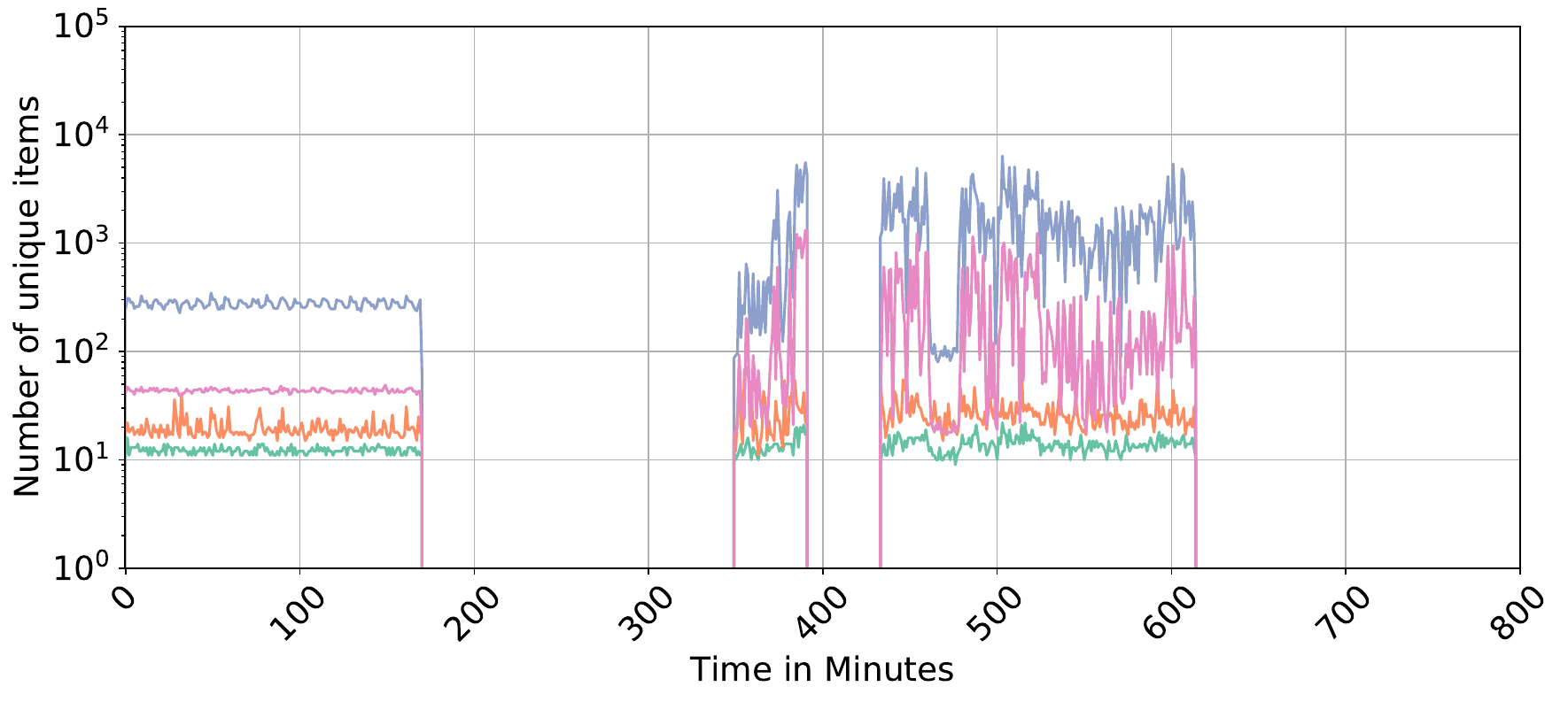}}\hspace{0.01cm}
    \subfloat[\centering][UNSW-NB15 Day 1]
    {\includegraphics[width=0.49\linewidth]{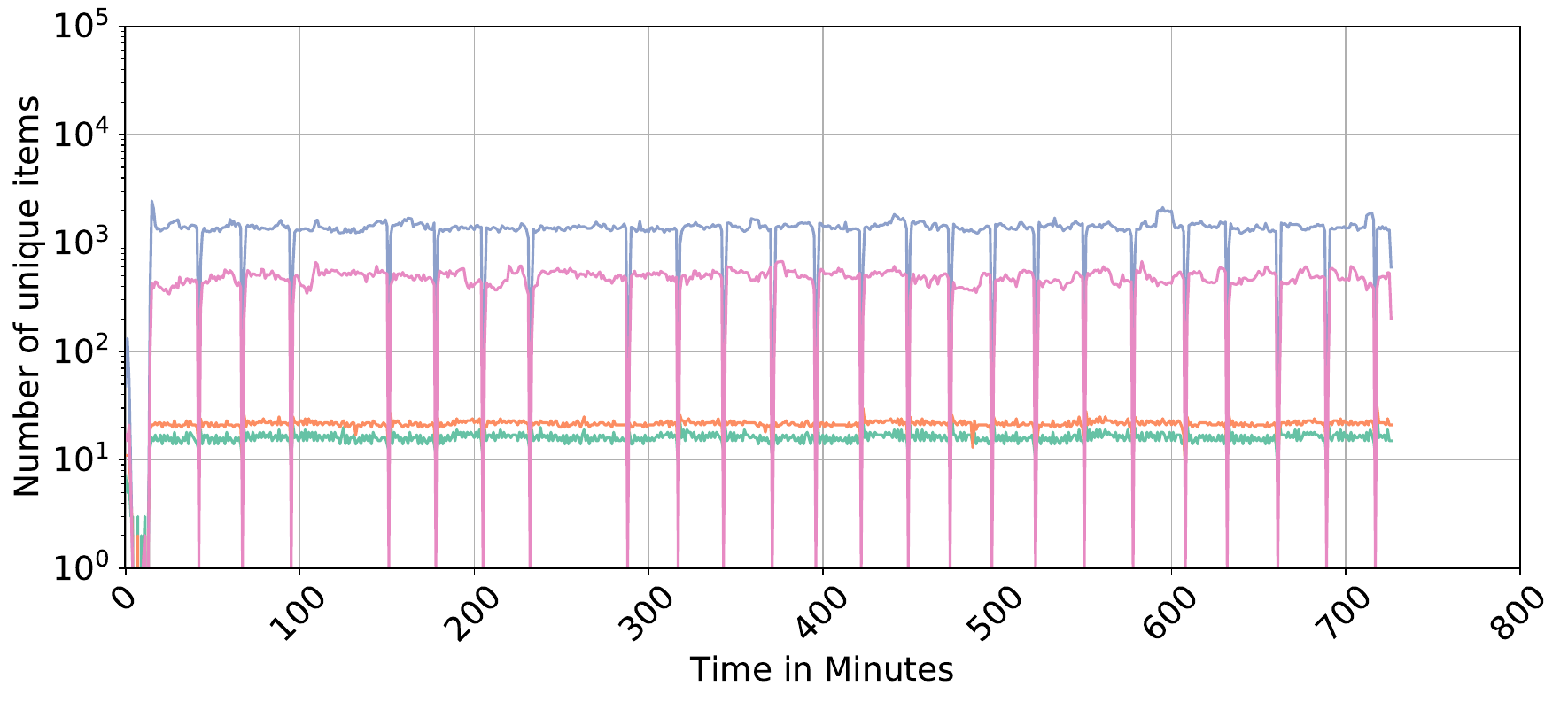}}
     \vspace{0.02cm} 
    \includegraphics[width=0.6\linewidth,trim=10 50 10 50,clip]{OCT_visualisation/Numerical/legend_only.pdf}
    \caption{Representation of categorical features in NF3-Datasets: IPV4\_SRC\_ADDR, IPV4\_DST\_ADDR, IPV4\_SRC\_PORT, and IPV4\_DST\_PORT. The x-axis represents time aggregated in minutes, and the y-axis shows the count of unique values for each category, highlighting the diversity in network activities over time.}
    \label{Categorical_TS}
\end{figure}

Similar to the previous subsection, Figure~\ref{Categorical_TS} visualises four categorical features: unique source and destination IP addresses and ports, captured in the same one-day snapshots. The x-axis represents time in minutes, while the y-axis shows the count of unique categorical values without repetition within each minute. Although the count is aggregated per minute, the data can be further zoomed in to monitor traffic at the level of seconds or even finer granularity. Here, we emphasise the utility of tracking categorical features over time, as it can assist in detecting certain types of anomalies related to source and destination IPs and ports.

In the NF3-CSE-CIC-IDS2018 Day 5, the count of unique source IPs (IPV4\_SRC\_ADDR) remains relatively steady, suggesting consistent activity from a stable set of source IPs throughout the day. Minor fluctuations in destination IPs (IPv4\_DST\_ADDR) may indicate interactions with a variety of external services or hosts. The source ports (L4\_SRC\_PORT) display stability with an occasional sharp spike, potentially pointing to a brief period of heightened network activity or an anomaly, while destination ports (L4\_DST\_PORT) show similar stability, suggesting regular communication patterns without significant anomalies. For NF3-ToN-IoT Day 5, both source and destination IPs exhibit peaks, notably in destination IPs, which could signify interactions with various external systems, potentially indicative of external data exchanges or scanning activities. Periodic spikes in both source and destination ports may indicate batched communications or network scans, suggesting an environment where network interactions are both dynamic and potentially vulnerable to security breaches.

The NF3-UNSW-NB15 Day 1 data reveals a low range of variation in both source and destination IPs, indicative of a controlled environment where a limited number of IPs are engaged. This suggests an environment with established, routine communication patterns, where ports show consistent levels, aligning with a network that experiences few irregularities and maintains a steady communication flow. In contrast, the NF3-BoT-IoT Day 1 plot maintains a lower count of unique source IPs with occasional spikes, suggesting sporadic activation of new source IPs possibly for command and control communications typical of a botnet scenario. Destination IPs show significant variability, likely related to the botnet's targets or a broader scope of victim engagement. The frequent changes in destination ports reflect dynamic interactions, potentially with multiple target machines or services, highlighting the erratic and potentially malicious nature of botnet activities within this dataset.

\subsection{Time-Frequency Representation}


Given the rich temporal information in network flows, various time and frequency signal processing techniques can be used for the analysis of the network traffic. 
Time-frequency analysis is a key signal processing technique that allows simultaneous examination of signals in both time and frequency domains, which can provide deeper insights into their underlying patterns.
This approach is particularly suited for non-stationary signals, where frequency content varies over time, such as in speech, music, and biomedical signals \cite{layeghy2014non,layeghy2014classification}.
Given the burstiness of network traffic~\cite{burstiness282603} where volumes can change rapidly (such as sudden spikes in packet volume during an attack) or exhibit periodicity (such as daily traffic patterns), it behaves as a time series signal with non-stationary properties~\cite{yang2023long}. Non-stationarity means the statistical properties, such as mean and variance, change over time; hence, conventional frequency domain approaches (like the Fourier transform) cannot deal with the time-varying and non-stationary nature of traffic pattern. Accordingly, time-frequency signal representation might be able to reveal patterns and anomalies in the time-frequency domain, which might be difficult to detect in the raw time-domain data.

\begin{figure}[!t]
    \centering
    \subfloat[\centering][DoS]
    {\includegraphics[width=0.32\linewidth]{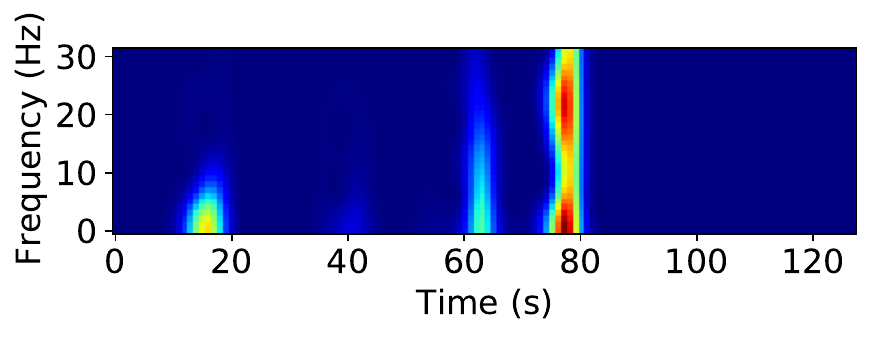}}\hspace{0.01cm}
    \subfloat[\centering][Backdoor]
    {\includegraphics[width=0.32\linewidth]{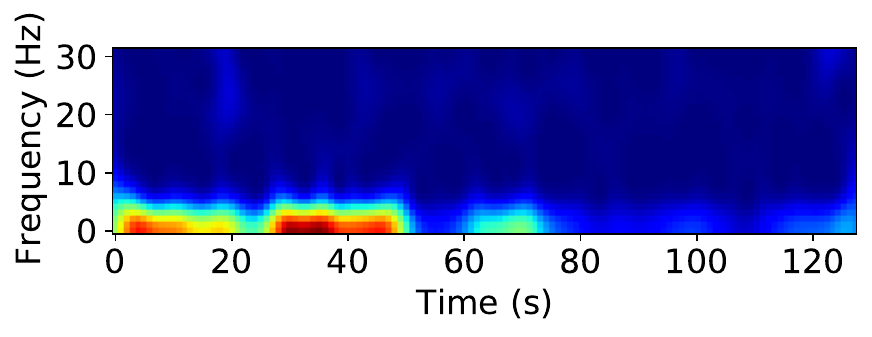}}\hspace{0.01cm}
    \subfloat[\centering][Analysis]
    {\includegraphics[width=0.32\linewidth]{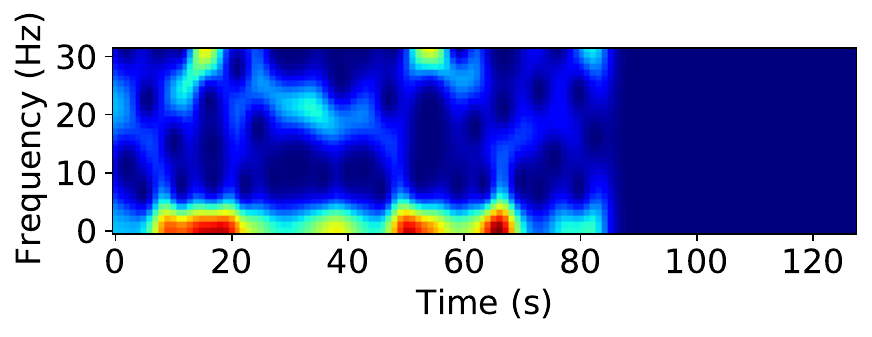}}
    
    \subfloat[\centering][Exploits]
    {\includegraphics[width=0.32\linewidth]{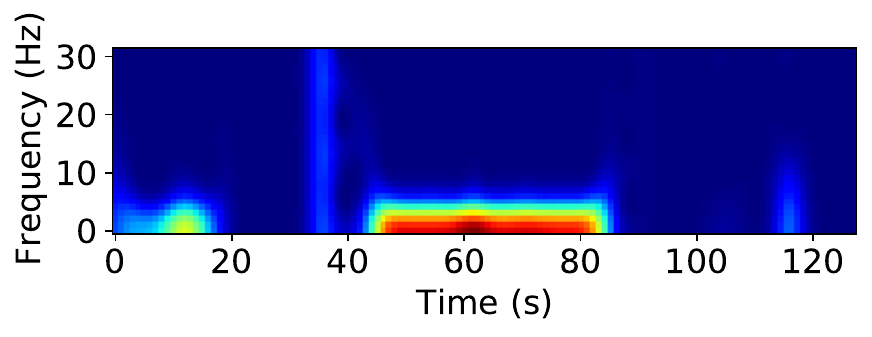}}\hspace{0.01cm}
    \subfloat[\centering][Fuzzers]
    {\includegraphics[width=0.32\linewidth]{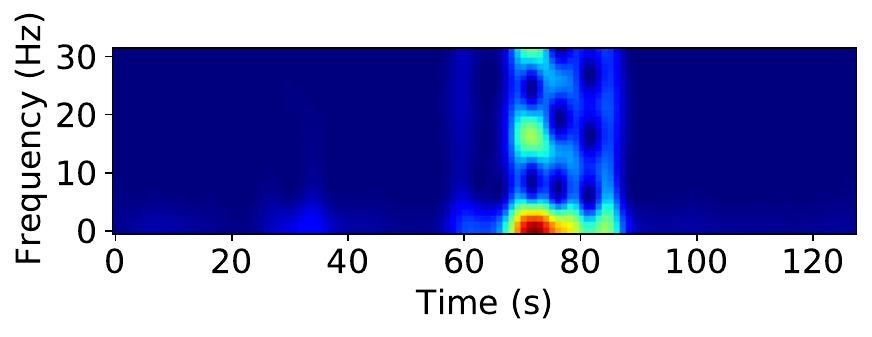}}\hspace{0.01cm}
    \subfloat[\centering][Generic]
    {\includegraphics[width=0.32\linewidth]{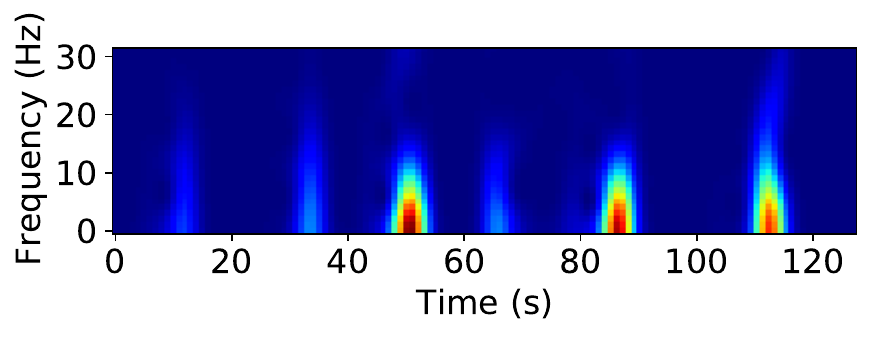}}
    
    \subfloat[\centering][Worms]
    {\includegraphics[width=0.32\linewidth]{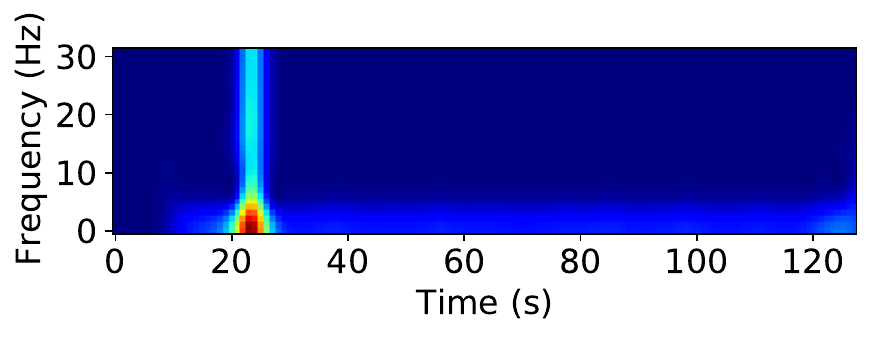}}\hspace{0.01cm}
    \subfloat[\centering][Shellcode]{\includegraphics[width=0.32\linewidth]{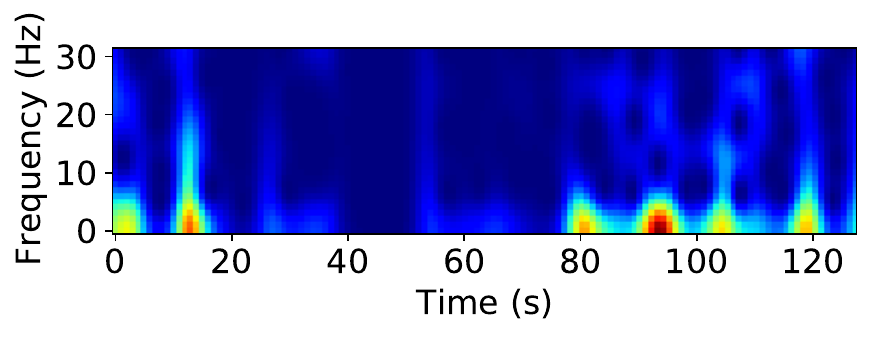}}\hspace{0.01cm}
    \subfloat[\centering][Reconnaissance]
    {\includegraphics[width=0.32\linewidth]{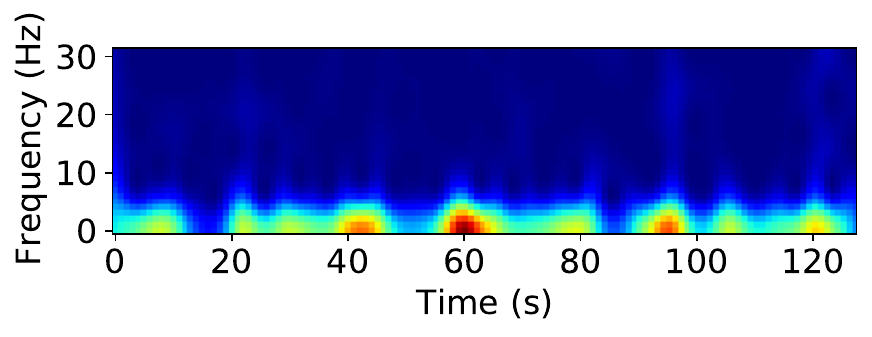}}
    \caption{Spectrogram representation of various attack classes of NF3-UNSW-NB15 dataset}
    \label{TFDs}
\end{figure}

Here, we explore one of these techniques, the spectrogram, to investigate the feasibility of such approaches in the field of ML-based NIDS. Spectrograms are the most common time-frequency techniques used to investigate signal variations over time. Using spectrograms, we can transform raw network flow time series into a richer representation that captures both frequency and temporal characteristics, potentially enhancing the performance of deep learning models.  
We focus on the NF3-UNSW-NB15 dataset. Figure~\ref{TFDs} shows the spectrogram of the most repeated pattern, for each attack class. 
As can be seen, the spectrogram of different classes vary significantly in some cases. For instance, while DoS and Worms share some similarities, their patterns still remain distinct from each other and from all other attack classes. Similarly, Fuzzers display a unique time-frequency signature, further differentiating them from other attack types.
These results highlight the potential of time-frequency representations in enhancing ML-based NIDS by providing a more detailed characterisation of network traffic patterns.



\section{Conclusion}

The increasing complexity of network traffic and diversity of modern attacks necessitates the incorporation of temporal analysis in network intrusion detection. Current attacks are no longer isolated events, but rather adaptive, time-evolving processes that can take advantage of timing vulnerabilities and encrypted traffic to evade detection. For instance, Advanced Persistent Threats (APTs) occur over extended periods of time, while low-and-slow attacks submerge malicious activity in normal traffic patterns. Additionally, the prevalence of encrypted protocols and the inadequacy of static analysis render temporal features (inter-packet arrival times, flow durations, traffic bursts) essential for detecting subtle attack behaviours. By analysing temporal dynamics, i.e. how the relationships and entities in a network change over time, researchers and practitioners can gain a deeper understanding of the evolving nature of network threats, enabling more effective detection and mitigation strategies.

In this paper, we try to address this gap by introducing a collection of four standardised NetFlow-based NIDS datasets enriched with detailed temporal features. Despite their importance, comprehensive temporal features have been largely absent from existing NetFlow-based NIDS datasets, limiting researchers' ability to study attack patterns over time across multiple datasets.
These datasets, the NF3 collection, provide a solid foundation for researchers and practitioners to dive into the temporal dynamics of network traffic. By incorporating precise flow start and end times, as well as detailed inter-packet arrival time statistics, these datasets provide a deeper understanding of attack patterns and network behaviour over time.

Our primary contribution in this study lies in conducting extensive temporal analysis to reveal the dynamics of network traffic and security threats. By visualising traffic distributions, flow length distributions by attack class, and time-frequency domain representations, this study has provided novel insights into network behaviour patterns.  
By making these temporal feature-enriched NetFlow datasets (NF3-Datasets) publicly available~\citep{uq_nids_datasets}, we aim to support ongoing research and development in ML-based network intrusion detection systems.
While this work highlights the importance of temporal features in NIDS, several challenges remain open for future exploration. Future research should focus on optimising ML models to leverage the temporal features introduced in this study effectively. Additionally, further work is needed to refine time-frequency-based approaches and evaluate their practicality in real-time intrusion detection scenarios. Investigating alternative temporal representations, such as recurrent neural networks (RNNs) and transformers, may also yield new insights into how sequential learning models can improve attack detection.

\bibliographystyle{unsrt} 
\bibliography{main}

@article{AI_IN_Cyber,
  title={Artificial intelligence and machine learning in cyber security},
  author={Prasad, Ramjee and Rohokale, Vandana and Prasad, Ramjee and Rohokale, Vandana},
  journal={Cyber security: the lifeline of information and communication technology},
  pages={231--247},
  year={2020},
  publisher={Springer}
}

@ARTICLE{CrossEvaluationML-NIDS,
  author={Apruzzese, Giovanni and Pajola, Luca and Conti, Mauro},
  journal={IEEE Transactions on Network and Service Management}, 
  title={The Cross-Evaluation of Machine Learning-Based Network Intrusion Detection Systems}, 
  year={2022},
  volume={19},
  number={4},
  pages={5152-5169},
  doi={10.1109/TNSM.2022.3157344}}

@Article{UnderstandNetworkBeforeML,
AUTHOR = {Tufail, Shahid and Riggs, Hugo and Tariq, Mohd and Sarwat, Arif I.},
TITLE = {Advancements and Challenges in Machine Learning: A Comprehensive Review of Models, Libraries, Applications, and Algorithms},
JOURNAL = {Electronics},
VOLUME = {12},
YEAR = {2023},
NUMBER = {8},
ARTICLE-NUMBER = {1789},
URL = {\href{https://www.mdpi.com/2079-9292/12/8/1789}},
ISSN = {2079-9292},
DOI = {10.3390/electronics12081789}
}

@misc{Temporal_Features_Importance,
      title={Meta-Analysis and Systematic Review for Anomaly Network Intrusion Detection Systems: Detection Methods, Dataset, Validation Methodology, and Challenges}, 
      author={Ziadoon K. Maseer and Robiah Yusof and Baidaa Al-Bander and Abdu Saif and Qusay Kanaan Kadhim},
      year={2023},
      eprint={2308.02805},
      archivePrefix={arXiv},
}

@article{SARHAN_SHAP,
title = {Evaluating Standard Feature Sets Towards Increased Generalisability and Explainability of ML-Based Network Intrusion Detection},
journal = {Big Data Research},
volume = {30},
pages = {100359},
year = {2022},
issn = {2214-5796},
doi = {10.1016/j.bdr.2022.100359},
url = {\href{https://www.sciencedirect.com/science/article/pii/S2214579622000533}},
author = {Mohanad Sarhan and Siamak Layeghy and Marius Portmann}
}

@article{benchmarkingThebenckmark,
title = {Benchmarking the benchmark — Comparing synthetic and real-world Network IDS datasets},
journal = {Journal of Information Security and Applications},
volume = {80},
pages = {103689},
year = {2024},
issn = {2214-2126},
doi = {10.1016/j.jisa.2023.103689},
url = {\href{https://www.sciencedirect.com/science/article/pii/S2214212623002739}},
author = {Siamak Layeghy and Marcus Gallagher and Marius Portmann}
}

@INPROCEEDINGS{Image-based-spec,
  author={Ahmad, Zeeshan and Khan, Adnan Shahid and Aqeel, Sehrish and Julaihi, Azlina Ahmadi and Tarmizi, Seleviawati and Annuar, Noralifah and Habeeb, Mohammed Sayeeduddin},
  booktitle={2022 Applied Informatics International Conference (AiIC)}, 
  title={S-ADS: Spectrogram Image-based Anomaly Detection System for IoT networks}, 
  year={2022},
  volume={},
  number={},
  pages={105-110},
  doi={10.1109/AiIC54368.2022.9914599}}

@ARTICLE{SlowDDoS_RelaredWork,
  author={Chen, Li Ming and Hsiao, Shun-Wen and Chen, Meng Chang and Liao, Wanjiun},
  journal={IEEE Systems Journal}, 
  title={Slow-Paced Persistent Network Attacks Analysis and Detection Using Spectrum Analysis}, 
  year={2016},
  volume={10},
  number={4},
  pages={1326-1337},
  doi={10.1109/JSYST.2014.2348567}}

@ARTICLE{Spectogram_Image,
  author={Khan, Adnan Shahid and Ahmad, Zeeshan and Abdullah, Johari and Ahmad, Farhan},
  journal={IEEE Access}, 
  title={A Spectrogram Image-Based Network Anomaly Detection System Using Deep Convolutional Neural Network}, 
  year={2021},
  volume={9},
  number={},
  pages={87079-87093},
  doi={10.1109/ACCESS.2021.3088149}}

@article{FLD_Analysis,
title = {Flow length and size distributions in campus Internet traffic},
journal = {Computer Communications},
volume = {167},
pages = {15-30},
year = {2021},
issn = {0140-3664},
doi = {10.1016/j.comcom.2020.12.016},
url = {\href{https://www.sciencedirect.com/science/article/pii/S0140366420320223}},
author = {Piotr Jurkiewicz and Grzegorz Rzym and Piotr Boryło}
}

@article{FLD_Classify,
  title={Classifying elephant and mice flows in high-speed scientific networks},
  author={Chhabra, Anshuman and Kiran, Mariam},
  journal={Proc. INDIS},
  pages={1--8},
  year={2017}
}

@inproceedings{netwrktrafficanomalydetection11,
  title={Network traffic characteristics of data centers in the wild},
  author={Benson, Theophilus and Akella, Aditya and Maltz, David A},
  booktitle={Proceedings of the 10th ACM SIGCOMM conference on Internet measurement},
  pages={267--280},
  year={2010}
}

@article{NIDS_Datasets_survey,
title = {A survey of network-based intrusion detection data sets},
journal = {Computers \& Security},
volume = {86},
pages = {147-167},
year = {2019},
issn = {0167-4048},
doi = {10.1016/j.cose.2019.06.005},
url = {\href{https://www.sciencedirect.com/science/article/pii/S016740481930118X}},
author = {Markus Ring and Sarah Wunderlich and Deniz Scheuring and Dieter Landes and Andreas Hotho},
}

@INPROCEEDINGS{KDD_limitations,
  author={Divekar, Abhishek and Parekh, Meet and Savla, Vaibhav and Mishra, Rudra and Shirole, Mahesh},
  booktitle={2018 IEEE 3rd International Conference on Computing, Communication and Security (ICCCS)}, 
  title={Benchmarking datasets for Anomaly-based Network Intrusion Detection: KDD CUP 99 alternatives}, 
  year={2018},
  volume={},
  number={},
  pages={1-8},
  doi={10.1109/CCCS.2018.8586840}}

@article{Dataset_Evaluation,
title = {A Review of the Advancement in Intrusion Detection Datasets},
journal = {Procedia Computer Science},
volume = {167},
pages = {636-645},
year = {2020},
note = {International Conference on Computational Intelligence and Data Science},
issn = {1877-0509},
doi = {10.1016/j.procs.2020.03.330},
url = {\href{https://www.sciencedirect.com/science/article/pii/S1877050920307961}},
author = {Ankit Thakkar and Ritika Lohiya}
}

@inproceedings{TheNatureOfDatacenter,
author = {Kandula, Srikanth and Sengupta, Sudipta and Greenberg, Albert and Patel, Parveen and Chaiken, Ronnie},
title = {The nature of data center traffic: measurements \& analysis},
year = {2009},
isbn = {9781605587714},
publisher = {Association for Computing Machinery},
address = {New York, NY, USA},
url = {\href{https://doi.org/10.1145/1644893.1644918}},
doi = {10.1145/1644893.1644918},
booktitle = {Proceedings of the 9th ACM SIGCOMM Conference on Internet Measurement},
pages = {202–208},
numpages = {7},
location = {Chicago, Illinois, USA},
series = {IMC '09}
}

@article{CIC2018_Original_dataset,
  title={Toward generating a new intrusion detection dataset and intrusion traffic characterization.},
  author={Sharafaldin, Iman and Lashkari, Arash Habibi and Ghorbani, Ali A and others},
  journal={ICISSp},
  volume={1},
  pages={108--116},
  year={2018}
}

@article{ToN_Original_dataset,
title = {A new distributed architecture for evaluating AI-based security systems at the edge: Network TON\_IoT datasets},
journal = {Sustainable Cities and Society},
volume = {72},
pages = {102994},
year = {2021},
issn = {2210-6707},
doi = {10.1016/j.scs.2021.102994},
url = {\href{https://www.sciencedirect.com/science/article/pii/S2210670721002808}},
author = {Nour Moustafa}
}

@article{BoT_Original_dataset,
title = {Towards the development of realistic botnet dataset in the Internet of Things for network forensic analytics: Bot-IoT dataset},
journal = {Future Generation Computer Systems},
volume = {100},
pages = {779-796},
year = {2019},
issn = {0167-739X},
doi = {10.1016/j.future.2019.05.041},
url = {\href{https://www.sciencedirect.com/science/article/pii/S0167739X18327687}},
author = {Nickolaos Koroniotis and Nour Moustafa and Elena Sitnikova and Benjamin Turnbull},
}

@INPROCEEDINGS{UNSW_Dataset_Original,
  author={Moustafa, Nour and Slay, Jill},
  booktitle={2015 Military Communications and Information Systems Conference (MilCIS)}, 
  title={UNSW-NB15: a comprehensive data set for network intrusion detection systems (UNSW-NB15 network data set)}, 
  year={2015},
  volume={},
  number={},
  pages={1-6},
  doi={10.1109/MilCIS.2015.7348942}}

@article{NF2_Datasets,
  title={Towards a standard feature set for network intrusion detection system datasets},
  author={Sarhan, Mohanad and Layeghy, Siamak and Portmann, Marius},
  journal={Mobile networks and applications},
  pages={1--14},
  year={2022},
  publisher={Springer}
}

@InProceedings{NF1_dataset,
author={Sarhan, Mohanad
and Layeghy, Siamak
and Moustafa, Nour
and Portmann, Marius},
editor={Deze, Zeng
and Huang, Huan
and Hou, Rui
and Rho, Seungmin
and Chilamkurti, Naveen},
title={NetFlow Datasets for Machine Learning-Based Network Intrusion Detection Systems},
booktitle={Big Data Technologies and Applications},
year={2021},
publisher={Springer International Publishing},
address={Cham},
pages={117--135},
isbn={978-3-030-72802-1}
}

@article{Niloo_Cattle,
title = {Deep learning-based cattle behaviour classification using joint time-frequency data representation},
journal = {Computers and Electronics in Agriculture},
volume = {187},
pages = {106241},
year = {2021},
issn = {0168-1699},
doi = {10.1016/j.compag.2021.106241},
url = {\href{https://www.sciencedirect.com/science/article/pii/S0168169921002581}},
author = {Seyedehfaezeh Hosseininoorbin and Siamak Layeghy and Brano Kusy and Raja Jurdak and Greg J. Bishop-Hurley and Paul L Greenwood and Marius Portmann}
}

@misc{Ntop,
  author = {Ntop},
  title = {nProbe, An Extensible NetFlow v5/v9/IPFIX Probe
for IPv4/v6},
  year = {2017},
  url = {\href{https://www.ntop.org/guides/nprobe/cli_options.html}},
  note = {Accessed: 2024-05-21}
}

@article{NIDS_Literature,
  title={A systematic literature review for network intrusion detection system (IDS)},
  author={Abdulganiyu, Oluwadamilare Harazeem and Ait Tchakoucht, Taha and Saheed, Yakub Kayode},
  journal={International Journal of Information Security},
  volume={22},
  number={5},
  pages={1125--1162},
  year={2023},
  publisher={Springer}
}

@article{Flow_Based_Literature,
title = {Flow-based intrusion detection: Techniques and challenges},
journal = {Computers \& Security},
volume = {70},
pages = {238-254},
year = {2017},
issn = {0167-4048},
doi = {10.1016/j.cose.2017.05.009},
url = {\href{https://www.sciencedirect.com/science/article/pii/S0167404817301165}},
author = {Muhammad Fahad Umer and Muhammad Sher and Yaxin Bi}
}

@inproceedings{input2024complexity,
  title={Measuring the complexity of benchmark NIDS datasets via spectral analysis},
  author={Flood, Robert and Aspinall, David},
  booktitle={2024 IEEE European Symposium on Security and Privacy Workshops (EuroS\&PW)},
  pages={335--341},
  year={2024},
  organization={IEEE}
}

@INPROCEEDINGS{stidm,
  author={Han, Xueying and Yin, Rongchao and Lu, Zhigang and Jiang, Bo and Liu, Yuling and Liu, Song and Wang, Chonghua and Li, Ning},
  booktitle={2020 IEEE 19th International Conference on Trust, Security and Privacy in Computing and Communications (TrustCom)}, 
  title={STIDM: A Spatial and Temporal Aware Intrusion Detection Model}, 
  year={2020},
  volume={},
  number={},
  pages={370-377},
  doi={10.1109/TrustCom50675.2020.00058}}

@ARTICLE{stidm2,
  author={Zhang, Yong and Chen, Xu and Jin, Lei and Wang, Xiaojuan and Guo, Da},
  journal={IEEE Access}, 
  title={Network Intrusion Detection: Based on Deep Hierarchical Network and Original Flow Data}, 
  year={2019},
  volume={7},
  number={},
  pages={37004-37016},
  doi={10.1109/ACCESS.2019.2905041}}

@article{zero_day_Attacks,
  title={A review of Machine Learning-based zero-day attack detection: Challenges and future directions},
  author={Guo, Yang},
  journal={Computer communications},
  volume={198},
  pages={175--185},
  year={2023},
  publisher={Elsevier}
}

@ARTICLE{Khumar2021ReviewNIDSTrends,
  author={Kumar, Satish and Gupta, Sunanda and Arora, Sakshi},
  journal={IEEE Access}, 
  title={Research Trends in Network-Based Intrusion Detection Systems: A Review}, 
  year={2021},
  volume={9},
  number={},
  pages={157761-157779},
  doi={10.1109/ACCESS.2021.3129775}}

@article{Mussa_Datasets_analysis,
  title={A detailed analysis of benchmark datasets for network intrusion detection system},
  author={Ghurab, Mossa and Gaphari, Ghaleb and Alshami, Faisal and Alshamy, Reem and Othman, Suad},
  journal={Asian Journal of Research in Computer Science},
  volume={7},
  number={4},
  pages={14--33},
  year={2021}
}

@INPROCEEDINGS{ahmed-survey,
  author={Ahmed, Lubna Ali Hassan and Hamad, Yahia Abdalla Mohamed and Abdalla, Ahmed Abdallah Mohamed Ali},
  booktitle={2022 International Arab Conference on Information Technology (ACIT)}, 
  title={Network-based Intrusion Detection Datasets: A Survey}, 
  year={2022},
  volume={},
  number={},
  pages={1-7},
  doi={10.1109/ACIT57182.2022.9994201}}

@inproceedings{empirical,
author = {Nychis, George and Sekar, Vyas and Andersen, David G. and Kim, Hyong and Zhang, Hui},
title = {An empirical evaluation of entropy-based traffic anomaly detection},
year = {2008},
isbn = {9781605583341},
publisher = {Association for Computing Machinery},
address = {New York, NY, USA},
url = {\href{https://doi.org/10.1145/1452520.1452539}},
doi = {10.1145/1452520.1452539},
booktitle = {Proceedings of the 8th ACM SIGCOMM Conference on Internet Measurement},
pages = {151–156},
numpages = {6},
location = {Vouliagmeni, Greece},
series = {IMC '08}
}

@misc{NetFlow_2004,
    series =    {Request for Comments},
    number =    3954,
    howpublished =  {RFC 3954},
    publisher = {RFC Editor},
    doi =       {10.17487/RFC3954},
    url =       {\href{https://www.rfc-editor.org/info/rfc3954}},
    author =    {Benoît Claise},
    title =     {{Cisco Systems NetFlow Services Export Version 9}},
    pagetotal = 33,
    year =      2004,
    month =     oct,
}

@INPROCEEDINGS{sok,
  author={Apruzzese, Giovanni and Laskov, Pavel and Schneider, Johannes},
  booktitle={2023 IEEE 8th European Symposium on Security and Privacy (EuroS\&P)}, 
  title={SoK: Pragmatic Assessment of Machine Learning for Network Intrusion Detection}, 
  year={2023},
  volume={},
  number={},
  pages={592-614},
  doi={10.1109/EuroSP57164.2023.00042}}

@inproceedings{structuralanalysis9,
author = {Lakhina, Anukool and Papagiannaki, Konstantina and Crovella, Mark and Diot, Christophe and Kolaczyk, Eric D. and Taft, Nina},
title = {Structural analysis of network traffic flows},
year = {2004},
isbn = {1581138733},
publisher = {Association for Computing Machinery},
address = {New York, NY, USA},
url = {\href{https://doi.org/10.1145/1005686.1005697}},
doi = {10.1145/1005686.1005697},
booktitle = {Proceedings of the Joint International Conference on Measurement and Modeling of Computer Systems},
pages = {61–72},
numpages = {12},
location = {New York, NY, USA},
series = {SIGMETRICS '04/Performance '04}
}

@article{mining_Anomalies2005,
author = {Lakhina, Anukool and Papagiannaki, Konstantina and Crovella, Mark and Diot, Christophe and Kolaczyk, Eric D. and Taft, Nina},
title = {Structural analysis of network traffic flows},
year = {2004},
issue_date = {June 2004},
publisher = {Association for Computing Machinery},
address = {New York, NY, USA},
volume = {32},
number = {1},
issn = {0163-5999},
url = {\href{https://doi.org/10.1145/1012888.1005697}},
doi = {10.1145/1012888.1005697},
journal = {SIGMETRICS Perform. Eval. Rev.},
month = jun,
pages = {61–72},
numpages = {12}
}

@article{expersknowledge,
title = {Effects of cyber security knowledge on attack detection},
journal = {Computers in Human Behavior},
volume = {48},
pages = {51-61},
year = {2015},
issn = {0747-5632},
doi = {10.1016/j.chb.2015.01.039},
url = {\href{https://www.sciencedirect.com/science/article/pii/S0747563215000539}},
author = {Noam Ben-Asher and Cleotilde Gonzalez}
}

@ARTICLE{elephant1,
  author={Hamdan, Mosab and Mohammed, Bushra and Humayun, Usman and Abdelaziz, Ahmed and Khan, Suleman and Ali, M. Akhtar and Imran, Muhammad and Marsono, M. N.},
  journal={IEEE Access}, 
  title={Flow-Aware Elephant Flow Detection for Software-Defined Networks}, 
  year={2020},
  volume={8},
  number={},
  pages={72585-72597},
  doi={10.1109/ACCESS.2020.2987977}}

@inproceedings{elephant2,
  title={An elephant flow detection method based on machine learning},
  author={Lou, Kaihao and Yang, Yongjian and Wang, Chuncai},
  booktitle={Smart Computing and Communication: 4th International Conference, SmartCom 2019, Birmingham, UK, October 11--13, 2019, Proceedings 4},
  pages={212--220},
  year={2019},
  organization={Springer}
}

@phdthesis{elephant3,
  title={Automatic detection of elephant flows through openflow-based openvswitch},
  author={Mallesh, Spurthi},
  year={2017},
  school={Dublin, National College of Ireland}
}

@inproceedings{roesch1999snort,
  title={Snort: Lightweight intrusion detection for networks.},
  author={Roesch, Martin and others},
  booktitle={Lisa},
  volume={99},
  pages={229--238},
  year={1999}
}

@ARTICLE{Review_ComputerVision,
  author={Zhao, Jiawei and Masood, Rahat and Seneviratne, Suranga},
  journal={IEEE Communications Surveys \& Tutorials}, 
  title={A Review of Computer Vision Methods in Network Security}, 
  year={2021},
  volume={23},
  number={3},
  pages={1838-1878},
  doi={10.1109/COMST.2021.3086475}}

@inproceedings{Temporal_NetFlow,
author = {Corsini, Andrea and Yang, Shanchieh Jay and Apruzzese, Giovanni},
title = {On the Evaluation of Sequential Machine Learning for Network Intrusion Detection},
year = {2021},
isbn = {9781450390514},
publisher = {Association for Computing Machinery},
address = {New York, NY, USA},
url = {\href{https://doi.org/10.1145/3465481.3470065}},
doi = {10.1145/3465481.3470065},
booktitle = {Proceedings of the 16th International Conference on Availability, Reliability and Security},
articleno = {113},
numpages = {10},
location = {Vienna, Austria},
series = {ARES '21}
}

@article{WhyFlowsDiffere,
  author={Vormayr, Gernot and Fabini, Joachim and Zseby, Tanja},
  journal={IEEE Communications Surveys \& Tutorials}, 
  title={Why are My Flows Different? A Tutorial on Flow Exporters}, 
  year={2020},
  volume={22},
  number={3},
  pages={2064-2103},
  doi={10.1109/COMST.2020.2989695}}

@inproceedings{NIDS_Anomaly_Review,
  title={Review on anomaly based network intrusion detection system},
  author={Samrin, Rafath and Vasumathi, D},
  booktitle={2017 international conference on electrical, electronics, communication, computer, and optimization techniques (ICEECCOT)},
  pages={141--147},
  year={2017},
  organization={IEEE}
}

@techreport{NetFlow,
  title={Cisco systems netflow services export version 9},
  author={Claise, Benoit},
  institution = {Cisco Systems},
  year={2004}
}

@article{harbic,
  title={HARBIC: Human activity recognition using bi-stream convolutional neural network with dual joint time--frequency representation},
  author={Hosseininoorbin, Seyedehfaezeh and Layeghy, Siamak and Kusy, Brano and Jurdak, Raja and Portmann, Marius},
  journal={Internet of Things},
  volume={22},
  pages={100816},
  year={2023},
  publisher={Elsevier}
}

@article{TPUNIDS,
  title={Exploring edge TPU for network intrusion detection in IoT},
  author={Hosseininoorbin, Seyedehfaezeh and Layeghy, Siamak and Sarhan, Mohanad and Jurdak, Raja and Portmann, Marius},
  journal={Journal of Parallel and Distributed Computing},
  volume={179},
  pages={104712},
  year={2023},
  publisher={Elsevier}
}

@article{manocchio2024flowtransformer,
  title={Flowtransformer: A transformer framework for flow-based network intrusion detection systems},
  author={Manocchio, Liam Daly and Layeghy, Siamak and Lo, Wai Weng and Kulatilleke, Gayan K and Sarhan, Mohanad and Portmann, Marius},
  journal={Expert Systems with Applications},
  volume={241},
  pages={122564},
  year={2024},
  publisher={Elsevier}
}

@inproceedings{layeghy2014non,
  title={{Non-invasive Monitoring of Fetal Movements Using Time-Frequency Features of Accelerometry}},
  author={Layeghy, Siamak and Azemi, Ghasem and Colditz, Paul and Boashash, Boualem},
  booktitle={2014 IEEE International Conference on Acoustics, Speech and Signal Processing (ICASSP)},
  pages={4379--4383},
  year={2014},
  organization={IEEE}
}

@inproceedings{layeghy2014classification,
  title={{Classification of Fetal Movement Accelerometry Through Time-Frequency Features}},
  author={Layeghy, Siamak and Azemi, Ghasem and Colditz, Paul and Boashash, Boualem},
  booktitle={2014 8th International Conference on Signal Processing and Communication Systems (ICSPCS)},
  pages={1--6},
  year={2014},
  organization={IEEE}
}

@article{yang2023long,
  title={{Long Term 5G Network Traffic Forecasting via Modeling Non-stationarity with Deep Learning}},
  author={Yang, Yuguang and Geng, Shupeng and Zhang, Baochang and Zhang, Juan and Wang, Zheng and Zhang, Yong and Doermann, David},
  journal={Communications Engineering},
  volume={2},
  number={1},
  pages={33},
  year={2023},
  publisher={Nature Publishing Group UK London}
}

@ARTICLE{burstiness282603,
  author={Leland, W.E. and Taqqu, M.S. and Willinger, W. and Wilson, D.V.},
  journal={IEEE/ACM Transactions on Networking}, 
  title={{On the Self-similar Nature of Ethernet Traffic }}, 
  year={1994},
  volume={2},
  number={1},
  pages={1-15},
  doi={10.1109/90.282603}}

@misc{uq_nids_datasets,
  author       = {Majed Luay and Siamak Layeghy and Seyedehfaezeh Hosseininoorbin and
Mohanad Sarhan and Nour Moustafa and Marius Portmann},
  title        = {{NetFlow V3 NIDS Datasets - The University of Queensland}},
  year         = {2025},
  note         = {Available at: \url{https://staff.itee.uq.edu.au/marius/NIDS_datasets/}}
}

\end{document}